%% file: TIP-12015-2014.tex
\documentclass[10pt,twocolumn,twoside]{IEEEtran}

\usepackage{graphicx}
\usepackage{caption}
\usepackage{array}
\usepackage{subfig}
\usepackage{cite}
\usepackage{stmaryrd}
\usepackage{amsfonts,latexsym,amssymb}
\usepackage[cmex10]{amsmath}
\usepackage{algorithm}
\usepackage{algorithmic}
\usepackage{hyperref}
\usepackage{multirow}
\usepackage{arydshln}

\input{zgx.math}

\newcommand{\nReal}{\Real_+}
\renewcommand{\fpartial}[2]{\ensuremath{\frac{\partial{#1}}{\partial{#2}}}}

\newtheorem{theorem}{{\bf{Proposition}}}

\newtheorem{corollary}{{\bf Corollary}}

\begin{document}
%
\title{Efficient Nonnegative Tucker Decompositions: Algorithms and Uniqueness}
{\author{Guoxu~Zhou, Andrzej~Cichocki ~\IEEEmembership{Fellow,~IEEE}, Qibin~Zhao, and Shengli~Xie \IEEEmembership{Senior Member,~IEEE,}

\thanks{Manuscript received ...This work was partially supported by the National Natural Science Foundation of China (grants U1201253),  the Guangdong Province Natural Science Foundation (2014A030308009), the Guangdong Province Excellent Thesis Foundation (SYBZZXM201316), and the JSPS KAKENHI (26730125, 15K15955).}
\thanks{ Guoxu Zhou is with the School of Automation at Guangdong University of Technology, Guangzhou, China                                 and  the Laboratory for Advanced Brain Signal Processing, RIKEN Brain Science Institute, Wako-shi, Saitama, Japan. E-mail: zhouguoxu@ieee.org.}%
\thanks{Andrzej Cichocki is with the Laboratory for Advanced Brain Signal Processing, RIKEN Brain Science Institute, Wako-shi, Saitama, Japan and with Systems Research Institute, Polish Academy of Science, Warsaw, Poland. E-mail: cia@brain.riken.jp.}%
\thanks{Qibin Zhao is with the Laboratory for Advanced Brain Signal Processing, RIKEN Brain Science Institute, Japan. E-mail: qbzhao@brain.riken.jp.}%
\thanks{Shengli Xie is with the Faculty of Automation, Guangdong University of Technology, Guangzhou 510006, China. E-mail: eeoshlxie@scut.edu.cn.}}
}

\markboth{IEEE TRANSACTIONS ON IMAGE PROCESSING}%
{ZHOU \MakeLowercase{\textit{et al.}}: Efficient Nonnegative Tucker Decompositions: Algorithms and Uniqueness}

\maketitle

\begin{abstract}
Nonnegative Tucker decomposition (NTD) is a powerful tool for the extraction of nonnegative parts-based and physically meaningful latent components from high-dimensional tensor data while preserving the natural multilinear structure of data. However, as the data tensor often has multiple modes and is large-scale, existing NTD algorithms suffer from a very high computational complexity in terms of both storage and computation time, which has been one major obstacle for practical applications of NTD. To overcome these disadvantages, we show how low (multilinear) rank approximation (LRA) of tensors is able to significantly simplify the computation of the gradients of the cost function, upon which a family of efficient first-order NTD algorithms are developed. Besides dramatically reducing the storage complexity and running time, the new algorithms are quite flexible and robust to noise because any well-established LRA approaches can be applied. We also show how nonnegativity incorporating sparsity substantially improves the uniqueness property and partially alleviates the curse of dimensionality of the Tucker decompositions. Simulation results on synthetic and real-world data justify the validity and high efficiency of the proposed NTD algorithms.

 \end{abstract}


%

\section{Introduction}
\label{sec:Intro}

\IEEEPARstart{F}{inding} information-rich and task-relevant variables hidden behind observation data is a fundamental task in data analysis and has been widely studied in the fields of signal and image processing and machine learning. Although the observation data can be very large, a much lower number of latent variables or components can capture the most significant features of the original data. By revealing such components, we achieve objectives such as dimensionality reduction and feature extraction and obtain a highly relevant and compact representation of high-dimensional data. This important topic has been extensively studied in the last several decades, particularly witnessed by the great success of blind source separation (BSS) techniques \cite{ComonBSS2010}. In these methods, observation data are modeled as a linear combination of the latent components that possess specific diversities such as statistical independence, a temporal structure, sparsity, and smoothness. By properly exploiting such diversities, a large family of matrix-factorization-based methodologies has been proposed and successfully applied to a variety of areas. In many applications, the data are more naturally represented by tensors, e.g., color images, video clips, and fMRI data. The methodologies that matricize the data and then apply matrix factorization approaches give a \emph{flattened view} of data and often cause a loss of the internal structure information; hence, it is more favorable to process such data in their own domain, i.e., tensor domain, to obtain a multiple perspective stereoscopic view of data rather than a flattened one. For this reason, tensor decomposition methods have been proposed and widely applied to deal with high-order tensors. As one of the most widely used methods, the Tucker decomposition has been applied to pattern recognition \cite{TuckerDA_TIP2007,TuckerDA_TPAMI2006}, clustering analysis \cite{NMF-book}, image denoising \cite{DenoisingHOSVD2013}, etc. and has achieved great success.

Very often, the observation data and latent components are naturally nonnegative, e.g., text, images, spectra, probability matrices, the adjacency matrices of graphs, web data based on impressions and clicks, and financial time series. For these data, the extracted components may lose most of their physical meaning if the nonnegativity is not preserved. In this regard, nonnegative matrix factorization (NMF) has been demonstrated to be a powerful tool to analyze nonnegative matrix data because NMF is able to give physically meaningful and more interpretable results. Particularly, it has the ability of learning the local parts of objects \cite{Lee1999}. As a result, NMF has received extensive study in the last decade \cite{NMF-book,Lee1999} and has been found many important applications including clustering analysis \cite{SPL_AONMF, TKDE2014_ZLi, PAMI2015_ZLi}, sparse coding \cite{sNMF2002},  dependent source separation \cite{nLCAIVM2010}, etc.  

\begin{figure*}[!t]
\centering
\includegraphics[width=.9\linewidth,height=0.25\linewidth]{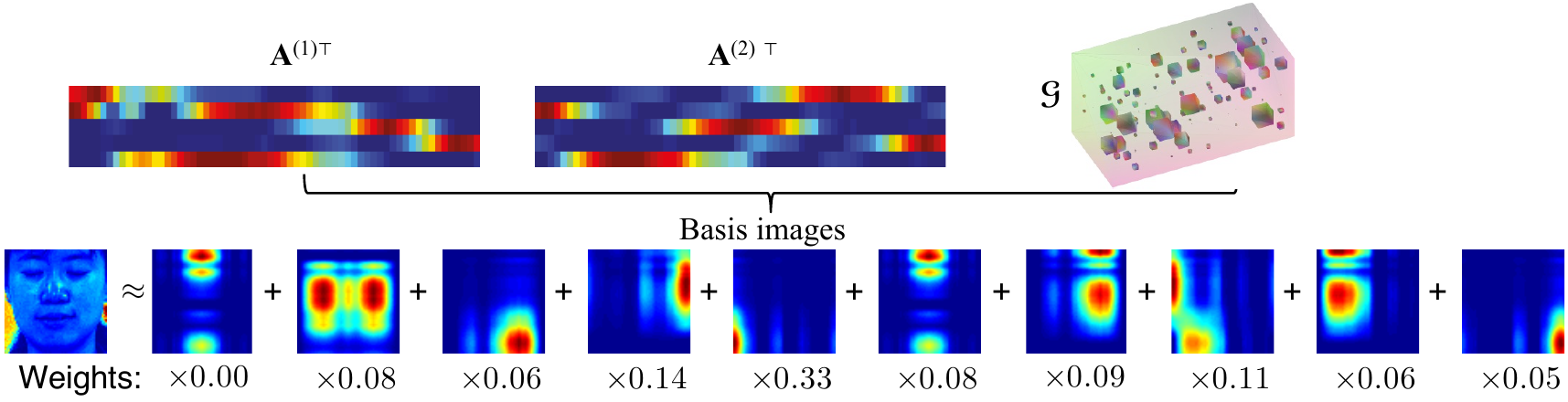}
\caption{An illustration of how NTD is able to give a local parts-based representation of tensor objects for the PIE face database. By NTD, each face is represented by a linear combination of a set of very sparse basis faces. In contrast to the basis images extracted by NMF, these basis faces possess a multilinear structure (i.e., $\tensor{G}\times_1\matn[1]{A}\times_2\matn[2]{A}$).}
\label{fig:NTDrep}
\end{figure*}

 \begin{table}[!t]
\caption{Notation and Definitions.}
\label{tab:notations}
\centerline{
\begin{tabular}{ >{\hfill}p{.1\textwidth} | p{0.35\textwidth}  }
\hline \hline
\mat{A}, \mats[r]{a}, $a_{ir}$ & A matrix, the $r$th-column, and the ($i,r$)th-entry of matrix \mat{A}, respectively. \\
 \mat{1}, \matO & The vector/matrix with all its elements being ones, zeros. \\
$\Set{I}_N$  & The index set of positive integers no larger than $N$ in ascending order, i.e., $\Set{I}_N=\set{1,2,\ldots,N}$. \\
$\nReal^{R_1\times R_2 \cdots \times R_N}$  & Set of $N$th-order nonnegative tensors (matrices) with the size of $R_1$-by-$R_2$$\cdots$by-$R_N$.  \\
$\mat{A}\ge\matO$  & Nonnegative matrix \mat{A}, i.e., $a_{ir}\ge0$, $\forall i,r$.   \\
$\nproj{\mat{A}}$ & An operator yielding a nonnegative matrix of $a_{ir}=\max(a_{ir},0)$, $\forall i,r$. \\
$\tensor{Y}$, \tenmat{Y} & A  tensor, the mode-$n$ matricization of tensor \tensor{Y}. \\
$\hdp$, $\matdiv{}{}$  & Element-wise (Hadamard) product, division  of matrices or tensors. Moreover, we define $\frac{\tensor{A}}{\tensor{B}}\defeq\matdiv{\tensor{A}}{\tensor{B}}$.\\
$\mat{C}=\mat{A}\kkp\mat{B}$ &  Kronecker product of $\mat{A}\in\Real^{I_1\times J_1}$ and $\mat{B}\in\Real^{I_2\times J_2}$ that yields  $\mat{C}=\left[a_{i_1j_1}\mat{B}\right]\in\Real^{I_1I_2\times J_1J_2}$ with entries $c_{(i_1-1)I_2+i_2,(j_1-1)J_2+j_2}=a_{i_1j_1}b_{i_2j_2}$.  \\
$\mat{C}=\mat{A}\krp\mat{B}$ 	& Khatri--Rao product of $\mat{A}=\begin{bmatrix}
\mats[1]{a} & \mats[2]{a} & \cdots & \mats[J]{a}
\end{bmatrix}\in\Real^{I_1\times J}$ and $\mat{B}=\begin{bmatrix}
\mats[1]{b} & \mats[2]{b} & \cdots & \mats[J]{b}
\end{bmatrix}\in\Real^{I_2\times J}$ that yields a matrix $\mat{C}=\begin{bmatrix}
\mats[1]{a}\kkp\mats[1]{b} & \mats[2]{a}\kkp\mats[2]{b} & \cdots & \mats[J]{a}\kkp\mats[J]{b}
\end{bmatrix}\in\Real^{I_1I_2\times J}$.  \\
$z_{\mat{A}}$, $s_{\mat{A}}$  & Number of zeros in $\mat{A}\in\Real^{I\times R}$, the sparsity defined as $s_\mat{A}\defeq{z_{\mat{A}}}/{(IR)}\in [0,1]$.  \\ 
$\mat{A}\sim U(0,1)$ & Elements of \mat{A} are drawn from independent uniform distributions between 0 and 1. \\
\hline \hline
\end{tabular}
}
\end{table}

For nonnegative tensor data analysis, nonnegative Tucker decomposition (NTD) has also gained  importance in recent years \cite{NMF-book,NTD-CVPR2007,HONMF:2008:Morup}. NTD not only inherits all of the advantages of NMF but also provides an additional multiway structured representation of data. \figurename \ref{fig:NTDrep} illustrates how NTD is able to give a parts-based representation of face images included in the PIE database\footnote{Available at \url{http://vasc.ri.cmu.edu/idb/html/face/}.}. In the figure, a sample face image is represented as a linear combination of a set of sparse basis images that possess a multilinear structure. Unconstrained Tucker decompositions are often criticized for the lack of uniqueness and the curse of dimensionality, which indicates that the size of the core tensor increases exponentially with the dimension. Compared with unconstrained Tucker decompositions, NTD is more likely to be unique and provides physically meaningful components. Moreover,  the core tensor in NTD is often very sparse, which allows us to discover the most significant links between components and to partially alleviate the curse of dimensionality.
Unfortunately,  existing algorithms  are generally performed by directly applying the NMF update rules without fully exploiting the special multilinear structure of the Tucker model, which in turn suffers from a very high computational complexity in terms of both space and time, especially when the data tensor is large-scale. It is therefore quite crucial to develop more efficient NTD algorithms that are able to yield satisfactory results within a tolerable time. By taking into account that unconstrained Tucker decompositions are significantly faster than NTD, we propose a new framework for efficient NTD that is based on an unconstrained Tucker decomposition of the  data tensor in this paper. As such, frequent access to the original big tensor is avoided, thereby leading to a considerably reduced computational complexity for NTD. Although the basic idea of NTD based on a proceeding low (multilinear) rank approximation (LRA) has been briefly introduced in our recent overview paper \cite{SPM_NMFNTD}, the detailed derivations are presented in this paper with new  results on the uniqueness of NTD. 

The rest of the paper is organized as follows. In Section II, the basic notation and NTD models are introduced. In Section III, the first-order NTD algorithms are reviewed. In  Section IV, flexible and efficient NTD algorithms  based on the low-rank approximation of data are introduced, and unique and sparse NTD is discussed in Section V. Simulations on synthetic and real-world data are presented in Section VI to demonstrate the high efficiency of the proposed algorithms, and conclusions are presented in Section VII.

The notation used in this paper is listed in TABLE \ref{tab:notations}, and more details can be found in \cite{Kolda09tensordecompositions, NMF-book}.

\section{NTD Models}
\label{sec:NTD} 
\subsection{Notation and Basic Multilinear Operators}
{\bf Definitions.} For an $N$th-order tensor $\tensor{G}\in\Real^{R_1\times R_2 \cdots\times R_N}$, we define
\begin{itemize}
\item {\bf Fibers.} A mode-$n$ fiber of tensor \tensor{G} is a vector obtained by fixing all indices but the $n$th index, e.g., $g_{r_1,\ldots,r_{n-1},:,r_{n+1},\ldots,r_N}$, by using the MATLAB colon operator.

\item {\bf Matricization.} The mode-$n$ matricization (unfolding) of \tensor{G} is an $R_n$-by-$\prod_{p\neq n}R_p$ matrix denoted by \tenmat{G} whose columns consist of all mode-$n$ fibers of \tensor{G}.

\item {\bf Mode-$n$ product.} The mode-$n$ product of \tensor{G} and an $I_n$-by-$R_n$ matrix $\matn{A}$ yields an $N$th-order tensor $\tensor{Y}=\tensor{G}\ttmn{A}\in\Real^{R_1\cdots\times R_{n-1}\times I_n \times R_{n+1}\cdots\times R_N}$ such that 
\begin{equation}
\notag
y_{r_1,\cdots,r_{n-1},i_n, r_{n+1}, \cdots,r_N}=\sum_{r_n=1}^{R_n}g_{r_1, r_2, \cdots, r_N}a_{i_n,r_n}.
\end{equation}
\end{itemize}

{\bf Useful properties.} The following properties concerning the mode-$n$ product will be frequently used:
\begin{enumerate}
\item If $\tensor{Y}=\tensor{G}\ttmn{A}$, then $\tenmat{Y}=\matn{A}\tenmat{G}$;
\item $(\tensor{G}\ttm{A})\ttm{B}=\tensor{G}\ttm{(BA)}$;
\item $(\tensor{G}\ttm[m]{A})\ttm[n]{B}=(\tensor{G}\ttm[n]{B})\ttm[m]{A}$, $n\neq m$.
\item If $\tensor{Y}=\tensor{G}\ttmn[1]{A}\ttmn[2]{A}\cdots\ttmn[N]{A}\defeq\tensor{G}\times_{n\in{\Set{I}_N}}\matn{A}$, then $\tenmat{Y}=\matn{A}\tenmat{G}\trans{\matn{R}{}}$, where 
\begin{equation}
\label{eq:kkp}
\begin{split}
\matn{R} & =\matn[N]{A}\kkp\cdots\kkp\matn[n+1]{A}\kkp\matn[n-1]{A}\kkp\cdots\kkp\matn[1]{A}\\
               & \defeq \bigkkp\nolimits_{{p\neq n}}\matn[p]{A},
\end{split}
\end{equation}
and \tenmat{Y} and \tenmat{G} are the mode-$n$ matricizations of \tensor{Y} and \tensor{G}, respectively.
\end{enumerate}
Note that owning to Property 3), the mode products in $\tensor{G}\times_{n\in{\Set{I}_N}}\matn{A}$ can be in any order of $n$. However, in $\bigkkp_{{p\neq n}}\matn[p]{A}$, the Kronecker products must be performed in the \emph{inverse} order of the index set $p\neq n\defeq\set{1,2,\ldots,n-1,n+1,\ldots,N}$.

\subsection{Nonnegative Tucker Decomposition Models}
\subsubsection{General NTD Model}
By Tucker decomposition, a given $N$th-order tensor $\tensor{Y}\in\Real^{I_1\times I_2 \cdots\times I_N}$ is approximated as
\begin{equation}
\label{eq:TuckerD}
\tensor{Y}\approx\tensor{G}\times_{n\in\Set{I}_N}\matn{A},
\end{equation}
where $\matn{A}=\begin{bmatrix}
\matn{a}_1 & \matn{a}_2 & \cdots & \matn{a}_{R_n}
\end{bmatrix}\in\Real^{I_n\times R_n}$ are the factor (component) matrices with $\rank{\matn{A}}=R_n$, and $\tensor{G}\in\Real^{R_1\times R_2\times \cdots\times R_N}$ is the core tensor whose entries reflect the interactions and connections between the components (columns) in different mode matrices.  We assume that $R_n\le I_n$  as high-dimensional data can often be well approximated by its lower-rank representations. 

In NTD, both the core tensor \tensor{G} and the factor matrices \matn{A} are required to be element-wisely nonnegative. The nonnegativity of the factors brings about two key effects: the resulting representation is purely additive but without subtractions, and the factors are often sparse, as they may contain many zero entries. These two effects equip nonnegative factorization methods with the ability of  learning localized parts of objects.

\subsubsection{Population NTD} In the above NTD, if we fix the $N$th factor matrix  $\matn[N]{A}$ to be the identity matrix $\matI$ (or equivalently, \matn[N]{A} is absorbed into the core tensor such that $\tensor{G}\from\tensor{G}\times_N\matn[N]{A}$ \cite{NMF-book}), we obtain the  population NTD model:
\begin{equation}
\label{eq:PNTD}
\tensor{Y}\approx \tensor{G}\times_{{n\neq N}}\matn{A}, \; \matn{A}\ge0, \;\tensor{G}\ge0.
\end{equation}
Population NTD  is important because it has a broad range of applications in machine learning and signal processing \cite{NMF-book}. To understand the key idea of population NTD, consider  simultaneously performing NTD on a set of $(N-1)$th-order sample tensors $\set{\tensor{Y}_i\in\Real_+^{I_1\times I_2 \cdots \times I_{N-1}}:i=1,2,\ldots,I_N}$  with \emph{common} component matrices. As such, each sample tensor can be represented as \cite{NMF-book}
\begin{equation}
\label{eq:PNTDrepresentation}
\tensor{Y}_i\approx\tensor{G}_i\times_{{n\neq N}}\matn{A}, \quad i=1,2,\ldots, I_N,
\end{equation}
where $\tensor{G}_i$ is the core tensor associated with $\tensor{Y}_i$, or equivalently,
\begin{equation}
\label{eq:PNTDMat}
\trans{\tenmat[N]{Y}}\approx\left[\bigkkp\nolimits_{{n\neq N}}\matn{A}\right]\trans{\tenmat[N]{G}},
\end{equation}
where each column of $\trans{\tenmat[N]{Y}}$ is a vectorized sample, and $\tenmat[N]{Y}$ is just the mode-$N$ matricization of the $N$th-order tensor \tensor{Y} obtained by  concatenating all of the samples $\tensor{Y}_i$. As such, all  samples are represented as a linear combination of $N-1$ sets of common basis vectors  (i.e., the columns of \matn{A}, ${n\neq N}$), and $\tensor{G}_i$ contains the extracted features.

In the case of $N=3$, the tensors $\tensor{Y}_i$ and $\tensor{G}_i$ in \eqref{eq:PNTDrepresentation} are  just matrices, and \eqref{eq:PNTDrepresentation}  can be written as
\begin{equation}
\label{eq:NTD_PVD}
\mat{Y}_i\approx\matn[1]{A}\mat{G}_i\trans{\matn[2]{A}{}},
\end{equation}
which has been studied in name of population value decomposition (PVD) \cite{PVDTucker} without nonnegativity constraints. 
 Hence, population NTD is an extension of PVD for extracting the nonnegative common components from multiblock higher-dimensional data equipped with the extra ability of learning the localized parts of objects \cite{PIEEE2015_LCA}.

An alternative method of performing such feature extraction tasks, which is referred to as nonnegative matrix factorization (NMF), vectorizes each sample to form a sample matrix $\mat{S}=\trans{\tenmat[N]{Y}}=\begin{bmatrix}
vec(\tensor{Y}_1) & vec(\tensor{Y}_2)& \cdots & vec(\tensor{Y}_{I_N})
\end{bmatrix}$. By using NMF,  \mat{S} is represented as
\begin{equation}
\label{eq:NMF}
\mat{S}=\trans{\tenmat[N]{Y}}\approx\mat{W}\trans{\mat{H}},\quad s.t. \;\; \mat{W}\ge\matO,\mat{H}\ge\matO,
\end{equation}
or equivalently, by using tensor notation,
\begin{equation}
\label{eq:TuckerNMF}
\mat{S}\approx\mat{W}\times_2\mat{H}, \quad s.t. \;\; \mat{W}\ge\matO,\mat{H}\ge\matO.
\end{equation}

An intuitive difference between population NTD and NMF is that the basis vectors in the former are the outer product of lower-dimensional vectors, as shown in \eqref{eq:PNTDMat}, which has a much lower number of free parameters and gives a kind of multilinear representation. This multilinear representation has been widely exploited to overcome the over-fitting problem in discriminant analysis \cite{TuckerDA_TPAMI2006, STDA_ERP2013}, and it substantially improves the sparsity of basis vectors, which will be discussed later.

\section{Overview of First-Order Methods for NTD}
In NTD, we need to minimize the following cost function:
\begin{equation}
\label{eq:NTDcost}
D_{\text{NTD}}=\frac{1}{2}\frob{\tensor{Y}-\tensor{\widehat{Y}}}^2,
\end{equation}
where $\tensor{\widehat{Y}}=\tensor{G}\times_{n\in\Set{I}_N}\matn{A}$ with the component matrices $\matn{A}\in\Real^{I_n\times R_n}_+$ and the core tensor $\tensor{G}\in\Real^{R_1\times R_2\cdots\times R_N}_+$, both of which are element-wisely nonnegative. 

Following the analysis in \cite{NTFComon2009} and similarly defining $\Delta^d=\set{\mat{x}\in\Real^{d+1}_+\|\frob[1]{\mat{x}}=1}$, we can straightforwardly obtain the following proposition:
\begin{theorem}
\label{th:Existance}
Let $\tensor{Y}\in\Real^{I_1\times I_2\cdots \times I_N}_+$. Then, the infimum
\begin{equation}
\notag
\begin{split}
\inf & \left\{ \frob[1]{\tensor{Y}-\tensor{G}\times_{n\in\Set{I}_N}\matn{A}} \right.\\
& \quad \left| \tensor{G}\in\Real_+^{R_1\times R_2\cdots\times R_N}, \matn{a}_j\in\Delta^{R_n-1}, j\in\Set{I}_{R_n}, n\in\Set{I}_N\right\}
\end{split}
\end{equation}
is attained.
\end{theorem}

Owning to Proposition \ref{th:Existance} and the equivalence of different norms, a global optimal solution for the problem in \eqref{eq:NTDcost} always exists. Below, we focus on the optimization algorithms.

To solve the optimization problem, we generally use a block coordinate descent framework: we minimize the cost function with respect to only the partial parameters (e.g., one factor matrix or even only one column of it) each time while fixing the others. To optimize \matn{A}, we consider an equivalent form of  \eqref{eq:NTDcost} by considering the mode-$n$ matricization of \tensor{Y} and $\tensor{\widehat{Y}}$:
\begin{equation}
\label{eq:NTDAn}
D_{\text{NTD}}=\frac{1}{2}\frob{\tenmat{Y}-\matn{A}\trans{\matn{B}{}}}^2, \; \matn{A}\ge\matO,
\end{equation}
where $\matn{B}=\matn{R}\trans{\tenmat{G}}$ and \matn{R} is defined as in \eqref{eq:kkp}. To optimize \tensor{G},  considering the vectorization of \tensor{Y} and \tensor{\widehat{Y}}, \eqref{eq:NTDcost} becomes 
\begin{equation}
\label{eq:NTDcore}
D_{\text{NTD}}=\frac{1}{2}\frob{\text{vec}(\tensor{Y})-\mat{F}\;\text{vec}(\tensor{G})}^2,\;\tensor{G}\ge\matO,
\end{equation}
where 
\begin{equation}
\label{eq:matF}
\mat{F}=\bigkkp\nolimits_{n\in\Set{I}_N}\matn{A}\in\Real^{(\prod_n I_n) \times (\prod_n R_n)}.
\end{equation}
Both \eqref{eq:NTDAn} and \eqref{eq:NTDcore} are nonnegative least squares (NLS) problems and have been extensively studied in the context of NMF, including the multiplicative update (MU) algorithm \cite{Lee2000}, the  hierarchical alternating least squares (HALS) method \cite{NMF-book}, the active-set methods \cite{kimHy:NMF_AS2008,NMF_BPP}, and the NMF algorithm based on the accelerated proximal gradient (APG)
 \cite{Guan2012}. These algorithms only use the first-order information and are free from searching the (learning) step size. To extend these methods to NTD, we need to compute the following respective gradients of $D_\text{NTD}$ with respect to \matn{A} and \tensor{G}:
\begin{equation}
\label{eq:Angrad}
\fpartial{D_\text{NTD}}{\matn{A}} = \matn{A}\trans{\matn{B}{}}\matn{B}-\tenmat{Y}\matn{B}, 
\end{equation}
where $\matn{{B}}=\left[\bigkkp_{{p\neq n}}\matn[p]{{A}}\right]\trans{\tenmat{G}}$, and
\begin{equation}
\label{eq:Ggradvec}
\fpartial{D_\text{NTD}}{\text{vec}(\tensor{G})} =\trans{\mat{F}}\mat{F}\;\text{vec}(\tensor{G})-\trans{\mat{F}}\;\text{vec}(\tensor{Y}),
\end{equation}
or equivalently,
\begin{equation}
\label{eq:Ggrad}
\fpartial{D_\text{NTD}}{\tensor{G}} = \tensor{\widehat{Y}}\times_{n\in\Set{I}_N}\trans{\matn{A}{}}-\tensor{Y}\times_{n\in\Set{I}_N}\trans{\matn{A}{}}.
\end{equation}
On the basis of \eqref{eq:Angrad}--\eqref{eq:Ggrad} and the existing NMF algorithms, a set of first-order NTD algorithms can be developed; for example, the NTD algorithms based on the MU and HALS algorithms have been developed in \cite{HONMF:2008:Morup,NTD-CVPR2007, HALSNTD:Phan2011}.

The above mentioned algorithms are all based on the gradients in \eqref{eq:Angrad}--\eqref{eq:Ggrad}. The major problem is that both \mat{F} and \matn{B} are quite large. For example, it can be verified that the complexity of computing $\tenmat{Y}\matn{B}$ is as high as \bigO{R_nI_1I_2\cdots I_N}. Hence, direct implementation of the above methods is extremely time and space demanding, especially for large-scale problems. 

\section{NTD Based on Low-Rank Approximations}
\label{sec:LRANTD} 
\subsection{Efficient Computation of Gradients by Using LRA}
To reduce the computational complexity,  we consider the following two-step approach to perform NTD:
\begin{enumerate}
\item {\bf the LRA step.} Obtain the LRA of \tensor{Y} such that 
\begin{equation}
\label{eq:HOSVD}
\tensor{Y}\approx\tensor{\widetilde{Y}}=\tensor{\widetilde{G}}\times_{n\in\Set{I}_N}\matn{\widetilde{A}},
\end{equation}
where $\matn{\widetilde{A}}\in\Real^{I_n\times\widetilde{R}_n}$, and $\widetilde{R}_n\ll I_n$ controls the approximation error and is not necessarily equal to $R_n$. 

\item {\bf the NTD step.} Perform NTD by minimizing $D_{\text{NTD}}=\frac{1}{2}\frob{\tensor{\widetilde{Y}}-\tensor{\widehat{Y}}}^2$, where $\tensor{\widehat{Y}}=\tensor{G}\times_{n\in\Set{I}_N}\matn{A}$ is the target nonnegative tensor.
\end{enumerate}
The effects of LRA are twofold: reduce the noise in the observation data and reduce the subsequent computational complexity in terms of both space and time. In fact, \tensor{\widetilde{Y}} consumes much less storage than \tensor{Y} does when $\widetilde{R}_n\ll I_n$: the former consumes $\sum_n(\widetilde{R}_nI_n)+\prod_n\widetilde{R}_n$, whereas \tensor{Y} consumes $\prod_nI_n$. For an intuitive comparison, suppose $N=4$, $R_n=10$, and $I_n=100$, $n=1,2,3,4$; then, the memory consumed by \tensor{\widetilde{Y}} is only 0.014\% of that consumed by \tensor{Y}.

Now, we show how the gradients with respect to \matn{A}, $n\in\Set{I}_N$, and \tensor{G} can be efficiently computed after \tensor{Y} is replaced with its low-rank approximation \tensor{\widetilde{Y}}.
First, we have
\begin{equation}
\label{eq:BtB}
\begin{split}
\trans{\matn{B}{}}\matn{B}  & = \tenmat{G}\trans{\matn{R}{}}\matn{R}\trans{\tenmat{G}} \\
& = \tenmat{G}\left[\bigkkp_{{p\neq n}}(\trans{\matn[p]{A}{}}\matn[p]{A})\right]\trans{\tenmat{G}}\\
&=\tenmat{X}\trans{\tenmat{G}},
\end{split}
\end{equation}
where \tenmat{X} is the mode-$n$ matricization of the tensor 
\begin{equation}
\label{eq:tenX}
\tensor{X}\defeq\tensor{G}\times_{{p\neq n}}(\trans{\matn[p]{A}{}}\matn[p]{A}).
\end{equation}
Here, \tensor{X} can be  efficiently computed  as it only involves the products of the small core tensor \tensor{G} with $R_n$-by-$R_n$ matrices.  Particularly, the memory-efficient (ME) tensor times the matrices proposed in \cite{Tucker_ME} can be applied to compute \tensor{X}  to further significantly reduce the memory consumption.

Regarding the term $\tenmat{Y}\matn{B}$, we have
\begin{equation}
\label{eq:YB}
\begin{split}
\tenmat{Y}\matn{B} 
& = \matn{\widetilde{A}}\tenmat{\widetilde{G}}\trans{\matn{\widetilde{R}}{}}\matn{R}\trans{\tenmat{G}} \\
&=\matn{\widetilde{A}}\tenmat{\widetilde{G}}\trans{\left(\bigkkp_{{p\neq n}}(\trans{\matn[p]{A}{}}\matn[p]{\widetilde{A}})\right)}\trans{\tenmat{G}} \\
&=\matn{\widetilde{A}}\tenmat{\widetilde{X}}\trans{\tenmat{G}},
\end{split}
\end{equation}
where \tenmat{\widetilde{X}} is the mode-$n$ matricization of the tensor
\begin{equation}
\label{eq:tildetenX}
\tensor{\widetilde{X}}=\tensor{\widetilde{G}}\times_{{p\neq n}}(\trans{\matn[p]{A}{}}\matn[p]{\widetilde{A}}).
\end{equation}

Furthermore, from \eqref{eq:tenX}, \eqref{eq:tildetenX}, and $\tensor{\widehat{Y}}=\tensor{G}\times_{n\in\Set{I}_N}\matn{A}$, we have  
\begin{equation}
\label{eq:FtF}
\tensor{\widehat{Y}}\times_{p\in\Set{I}_N}\trans{\matn[p]{A}{}}=\tensor{X}\times_n(\trans{\matn{A}{}}\matn{A}),
\end{equation}
and
\begin{equation}
\label{eq:YttmA}
\tensor{Y}\times_{p\in\Set{I}_N}\trans{\matn[p]{A}{}}=\tensor{\widetilde{X}}\times_n(\trans{\matn{{A}}{}}\matn{\widetilde{A}}).
\end{equation}
The tensors \tensor{X} and \tensor{\tilde{X}} can be computed very efficiently, as they only involve  multiplications between very small core tensors and matrices. On the basis of the above analysis, the gradients \eqref{eq:Angrad} and \eqref{eq:Ggrad} can be computed as
\begin{equation}
\label{eq:LRAgrad}
\begin{split}
\fpartial{D_\text{NTD}}{\matn{A}} &= \matn{A}\left(\tenmat{X}\trans{\tenmat{G}}\right)-\matn{\widetilde{A}}\left(\tenmat{\widetilde{X}}\trans{\tenmat{G}}\right), \\
\fpartial{D_\text{NTD}}{\tensor{G}} &= \tensor{X}\times_n(\trans{\matn{A}{}}\matn{A})-\tensor{\widetilde{X}}\times_n(\trans{\matn{{A}}{}}\matn{\widetilde{A}}).
\end{split}
\end{equation}

We counter the floating-point multiplications to measure the computational complexity on the condition that $\widetilde{R}_n=R_n=R$ and $I_n=I$, $\forall n\in\Set{I}_N$, from which we can see that the LRA versions are significantly faster than the original versions. 

\renewcommand{\arraystretch}{1.8}
 \begin{table}[!t]
\caption{Floating-point Multiplication Counts for the Computation of the Gradients under the Conditions That $R_n=\tilde{R}_n=R$ and $I_n=I$, $n\in\Set{I}_N$.}
\label{tab:flipMult}
\centerline{
\begin{tabular}{ r | c | c  }
\hline \hline
&   With LRA &  Without LRA \\ \hline
$\fpartial{D_\text{NTD}}{\matn{A}}$ & $2IR^2+2R^{N+1}$ & $IR^2+R^{N+1}+\sum\limits_{k=1}^NR^kI^{N+1-k}$ \\ \hline
\fpartial{D_\text{NTD}}{\tensor{G}}  &  $2IR^2+2R^{N+1}$ & $IR^2+R^{N+1}+\sum\limits_{k=1}^NR^kI^{N+1-k}$ \\
\hline  \hline
\end{tabular}
}
\end{table}
\renewcommand{\arraystretch}{1}

Low-rank approximation  or unconstrained Tucker decomposition  of the data tensor  plays a key role in the proposed two-step framework.  The high-order singular value decomposition (HOSVD) method \cite{HOSVD2000}  often serves as the workhorse for this purpose. Although it provides a good trade-off between  accuracy and efficiency,   it involves the eigenvalue decomposition of the very large matrices \tenmat{Y};  hence, it is not suitable for very large-scale problems \cite{paracomp}. Its memory efficient variant proposed in \cite{Tucker_ME} is an iterative method and provides improved scalability.  CUR decomposition \cite{CURTensorCaiafaC2010} and the MACH method \cite{MACH2010}, which are respectively based on sampling and sparsification, provide favorable scalability and are suitable for  large-scale problems. In \cite{bigTensor2014}, a highly scalable Tucker decomposition algorithm was  developed  on the basis of distributed computing and randomization. All of these methods have assumed that the noise is Gaussian. Otherwise, robust tensor decomposition methods \cite{RTucker:SIAM2014, zhao2015bayesian} are recommended if the  data tensor is contaminated by outliers. In practice, which one is the best depends on many factors, e.g., whether the data is  sparse or dense, the scale of data,  the noise distribution, etc.

\subsection{Efficient NTD Algorithms Based on LRA}

\subsubsection{LRANTD Based on MU rules (MU-NTD)} the standard MU method updates \matn{A}  using
\begin{equation}
\label{eq:MU}
\matn{A}\from\matn{A}-\boldsymbol{\eta}_{\matn{A}}\hdp\fpartial{D_\text{NTD}}{\matn{A}}
\end{equation}
with a clever choice of step size $\boldsymbol{\eta}_{\matn{A}}=\matdiv{\matn{A}}{(\matn{A}\tenmat{X}\trans{\tenmat{G}})}$. As such, the cost function remains nonincreasing and  \matn{A} remains nonnegative. As the term $\matn{\widetilde{A}}\tenmat{\widetilde{X}}\trans{\tenmat{G}}$ (i.e., \eqref{eq:YB}) may contain some negative elements after LRA, we apply the method proposed in \cite{TSP-lraNMF} to \eqref{eq:MU}, where the descent direction $\fpartial{D_\text{NTD}}{\matn{A}}$ is replaced by $\matn{A}\tenmat{X}\trans{\tenmat{G}}-\proj_+(\matn{\widetilde{A}}\tenmat{\widetilde{X}}\trans{\tenmat{G}})$, thereby leading to the following MU  formula:
\begin{equation}
\label{eq:MULRA_A}
\matn{A} \from \matn{A}\hdp\frac{\proj_+\left(\matn{\widetilde{A}}\tenmat{\widetilde{X}}\trans{\tenmat{G}}\right)}{\matn{A}\tenmat{X}\trans{\tenmat{G}}}.
\end{equation}
Similarly, we obtain the MU  rule for  \tensor{G} as
\begin{equation}
\label{eq:MULRA_G}
\tensor{G}  \from \tensor{G}\hdp \frac{\proj_+\left(\tensor{\widetilde{X}}\times_n(\trans{\matn{{A}}{}}\matn{\widetilde{A}})\right)}{(\tensor{X}\times_n(\trans{\matn{{A}}{}}\matn{A})}.
\end{equation}
See Algorithm \ref{alg:MU-NTD} for the pseudocode of the NTD algorithm based on the MU rules in\eqref{eq:MULRA_A} and \eqref{eq:MULRA_G}. (They are quite different from the algorithms proposed in \cite{HONMF:2008:Morup,NTD-CVPR2007,HALSNTD:Phan2011} that update the parameters in a very inefficient manner: they update one parameter only once in one main iteration. In our case, however, multiple updates will be used to achieve a sufficient decrease of the cost function to improve the total efficiency, motivated by the work \cite{accNMF_Gillis} for NMF. Of course, in order to achieve a high efficiency,  exact convergence of each subproblem is generally unnecessary during the iterations.)

\renewcommand{\matdiv}[2]{\frac{#1}{#2}}
\begin{algorithm}
\caption{The MU-NTD Algorithm}
\label{alg:MU-NTD}
\begin{algorithmic}[1]
\REQUIRE \tensor{Y}.
\STATE Initialization.
\STATE Perform LRA: $\tensor{Y}\from\tensor{\widetilde{G}}\times_{n\in\Set{I}_N}\matn{\widetilde{A}}$.
\WHILE{not converged}
\STATE Execute the loop in lines 5--8 for $n=1,2,\ldots,N$:
\WHILE {not converged}
\STATE Update the tensors \tensor{X} and \tensor{\widetilde{X}} using \eqref{eq:tenX} and \eqref{eq:tildetenX}.
\STATE $\matn{A} \from \matn{A}\hdp\matdiv{\mathcal{P}_+(\matn{\widetilde{A}}\tenmat{\widetilde{X}}\trans{\tenmat{G}})}{(\matn{A}\tenmat{X}\trans{\tenmat{G}})}$.
\ENDWHILE
\WHILE {not converged}
\STATE $\tensor{G}  \from \tensor{G}\hdp \matdiv{\proj_+(\tensor{\widetilde{X}}\times_n(\trans{\matn{{A}}{}}\matn{\widetilde{A}}))}{(\tensor{X}\times_n(\trans{\matn{{A}}{}}\matn{A})}$.
\ENDWHILE
\ENDWHILE
\RETURN \tensor{G}, \matn{A}, $n\in\Set{I}_N$.
\end{algorithmic}
\end{algorithm}

\subsubsection{NTD Based on  HALS (HALS-NTD)}  HALS-NTD updates only one column of \matn{A} each time \cite{HALSNTD:Phan2011} by minimizing
\begin{equation}
\label{eq:HALScost}
D_\text{NTD}=\frac{1}{2}\frob{\tenmat{Y}^r-\matn{a}_r\trans{\matn{b}_r{}}}^2,
\end{equation}
where $\tenmat{Y}^r=\tenmat{Y}-\sum_{i\neq r}\matn{a}_i\trans{\matn{b}_i{}}$,  $r\in\Set{I}_{R_n}=\set{1,2,\ldots,R_n}$.
By using the Lagrange multiplier method \cite{TSP-lraNMF}, we obtain the update rule for $\matn{a}_r$:
\begin{equation}
\label{eq:HALS}
\matn{a}_r \from \matn{a}_r+\frac{1}{t^{(n)}_{r,r}}\proj_+{\left(\matn{q}_r-\matn{A}\matn{t}_r\right)},
\end{equation}
where $t^{(n)}_{r,r}$ is the $(r,r)$ entry of the matrix \matn{T}; $\matn{t}_r$ and $\matn{q}_r$ are the $r$th columns of the matrices \matn{T} and \matn{Q}, respectively;  $\matn{Q}=\matn{\widetilde{A}}\tenmat{\widetilde{X}}\trans{\tenmat{G}}$; and $\matn{T}=\tenmat{X}\trans{\tenmat{G}}$. The MU rule in \eqref{eq:MULRA_G} can be used to update  \tensor{G}.

\subsubsection{NTD Based on the Accelerated Proximal Gradient  (APG-NTD)} Following the analysis in \cite{Guan2012}, we have the following:
\begin{theorem}
\label{th:Lipschitz}
Both \fpartial{D_\text{NTD}}{\matn{A}} and \fpartial{D_\text{NTD}}{\tensor{G}} in  \eqref{eq:Angrad}, \eqref{eq:Ggrad} and \eqref{eq:LRAgrad} are Lipschitz continuous with the Lipschitz constants $L_{\matn{A}}=\frob[2]{\matn{T}}$ and $L_{\tensor{G}}=\frob[2]{\trans{\mat{F}}\mat{F}}=\frob[2]{\prod_{n\in\Set{I}_N}\trans{\matn{A}{}}\matn{A}}$. 
\end{theorem}

Hence, the APG method can be applied to update \matn{A} and \tensor{G}. For example, we use  Algorithm \ref{alg:APG} to obtain \tensor{G} provided that all \matn{A} are fixed. Similarly, we can obtained the update rules for \matn{A}. We call this algorithm APG-NTD; see Algorithm \ref{alg:APG}.

\begin{algorithm}
\caption{The APG Algorithm for the Core Tensor \tensor{G}}
\label{alg:APG}
\begin{algorithmic}[1]
\REQUIRE \tensor{Y}, \matn{A}.
\STATE Initialize $\tensor{E}_1=\tensor{G}_0$, $\alpha_0=1$, $L_{\tensor{G}}=\frob{\trans{\mat{F}}\mat{F}}$, $k=1$.
\WHILE {not converged}
\STATE $\tensor{G}_k \from \proj_+\left(\tensor{E}_k-\frac{1}{L_{\tensor{G}}}\fpartial{D_\text{NTD}}{\tensor{G}}\right)$,
\STATE $\alpha_{k+1} \from \frac{1+\sqrt{4\alpha_k^2+1}}{2}$,
\STATE $\tensor{E}_{k+1} \from \tensor{G}_k+\frac{\alpha_{k}-1}{\alpha_{k+1}}(\tensor{G}_k-\tensor{G}_{k-1})$,
\STATE $k\from k+1$.
\ENDWHILE
\RETURN \tensor{G}.
\end{algorithmic}
\end{algorithm}


\subsubsection{Active Set Method (AS-NTD)} The active set method proposed for NMF in \cite{kimHy:NMF_AS2008,NMF_BPP} can be applied to NTD  to solve \eqref{eq:NTDAn} and \eqref{eq:NTDcore}. Roughly speaking, these methods involve solving the inverse problems of $\fpartial{D_\text{NTD}}{\text{vec}(\matn{A})}=\matO$ and $\fpartial{D_\text{NTD}}{\text{vec}(\tensor{G})}=\matO$ under nonnegativity constraints, and among them,  block principal pivoting (BPP) achieved the best performance, as multiple columns are updated simultaneously \cite{NMF_BPP}. Active-set-based NMF approaches converge very fast, but their stability was questioned in some cases \cite{Guan2012}.

\subsubsection{Alternating Least Squares (ALS) and Semi-NTD} 
Sometimes some component matrices \matn{A} and/or the core tensor \tensor{G} are not necessarily nonnegative, which is a natural extension of semi-NMF \cite{semiNMF:PAMI2010}. These factors can be updated by their least-squares (LS) solutions to the linear equation systems $\fpartial{D_\text{NTD}}{\matn{A}}=\matO$ and $\fpartial{D_\text{NTD}}{\tensor{G}}=\matO$, which results in
\begin{equation}
\label{eq:ALSTucker}
\begin{split}
\matn{A} &\from \matn{\widetilde{A}}\tenmat{\widetilde{X}}\trans{\tenmat{G}}\left(\tenmat{X}\trans{\tenmat{G}}\right)^{-1},\\
\tensor{G} & \from \tensor{G}\times_{n\in\Set{I}_N}[{(\trans{\matn{A}{}}\matn{A})}^{-1}\trans{\matn{A}{}}\matn{\widetilde{A}}].
\end{split}
\end{equation}
Note that if we apply an additional nonnegativity projection to \eqref{eq:ALSTucker}, a very simple ALS-based NTD algorithm (ALS-NTD) yields
\begin{equation}
\label{eq:ALSNTD}
\matn{A}=\proj_+(\matn{A}),\quad \tensor{G}=\proj_+(\tensor{G}).
\end{equation}
Similar to the ALS-based NMF method,  the ALS-NTD method generally has no guarantee of convergence. However, many experimental results show that this method works quite well when $R_n\ll I_n$, $\forall n\in\Set{I}_N$, and the factors are very sparse.

\subsection{Error Bounds}
An important question is how the LRA will affect the accuracy of NTD.  The following proposition provides an approximate error bound:
\begin{theorem}
\label{th:ErrB}
Let tensor $\tensor{\widetilde{Y}}=\tensor{\widetilde{G}}\times_{n\in\Set{I}_N}\matn{\widetilde{A}}$ be a low-rank approximation of \tensor{Y} with $\frob{\tensor{Y}-\tensor{\widetilde{Y}}}\le\sigma$. Suppose that  $\tensor{\widehat{Y}}=\tensor{\widehat{G}}\times_{n\in\Set{I}_N}\matn{\widehat{A}}$ and $\tensor{\acute{Y}}=\tensor{\acute{G}}\times_{n\in\Set{I}_N}\matn{\acute{A}}$ are the optimal NTDs of \tensor{\widetilde{Y}} and \tensor{Y}, respectively, and $\frob{\tensor{Y}-\tensor{\acute{Y}}}=\epsilon$. Then 
\begin{equation}
\epsilon\le\frob{\tensor{Y}-\tensor{\widehat{Y}}}\le 2\sigma+\epsilon.
\end{equation}
\end{theorem}
\begin{IEEEproof}
As $\tensor{\widehat{Y}}$ and $\tensor{\acute{Y}}$ are respectively the optimal NTDs of $\tensor{\widetilde{Y}}$ and \tensor{Y}, we have
\begin{equation}
\begin{split}
\epsilon & \le\frob{\tensor{Y}-\tensor{\widehat{Y}}} 
 \le \frob{\tensor{Y}-\tensor{\widetilde{Y}}}+\frob{\tensor{\widetilde{Y}}-\tensor{\widehat{Y}}}\\
& \le \sigma+\frob{\tensor{\widetilde{Y}}-\tensor{\acute{Y}}}\\
& \le \sigma+\frob{\tensor{\widetilde{Y}}-\tensor{{Y}}}+\frob{\tensor{{Y}}-\tensor{\acute{Y}}}\\
& \le 2\sigma+\epsilon
\end{split}
\end{equation}
\end{IEEEproof}
Obviously, if the LRA is exact such that $\sigma=0$, there is no difference between the direct NTD and that based on LRA. In summary, the quality of LRA could be crucial to achieve satisfactory accuracy.

\subsection{NTD with Missing Values (Weighted NTD)}
In practice, some entries of the data tensor could be severely contaminated by noise and hence could not be used or they are simply missing. In such a case, the intrinsic low-rank structure of data often allows the recovery of the missing values by using incomplete data.  NTD with missing values can be formulated as the following general weighted NTD problem:
\begin{equation}
\label{eq:MissingNTD}
\min_{\tensor{G}\ge\matO,\;\matn{A}\ge\matO}\quad \frob{\tensor{W}\hdp\left(\tensor{Y}-\tensor{G}\times_{n\in\Set{I}_N}\matn{A}\right)},
\end{equation}
where the entries of the weight tensor \tensor{W} are between 0 and 1.  If the entries of \tensor{W} can only be either 0 or 1, \eqref{eq:MissingNTD} is the problem of NTD with missing values. Although there have been many methods proposed for tensor/matrix decompositions with missing values that can be straightforwardly extended to NTD, an ad-hoc two-step solution can be applied: in Step 1, weighted Tucker decomposition is performed by minimizing the cost function $\frob{\tensor{W}\hdp\left(\tensor{Y}-\tensor{\widetilde{Y}}\right)}$, where $\tensor{\widetilde{Y}}=\tensor{G}\times_{n\in\Set{I}_N}\matn{A}$; then, in Step 2, the NTD is performed by using the completed tensor \tensor{\widetilde{Y}} yielded in Step 1.

Notice that weighted Tucker decomposition approaches also allow us to obtain the  low-rank approximations  by accessing only randomly sampled entries (fibers) of a high-dimensional tensor, which is a very useful technique to deal with large-scale problems \cite{CURTensorCaiafaC2010}. Although all of the approaches proposed for the missing-values problem and those based on random sampling attempt to find the optimal approximation to the original data by using only partial data, they have a subtle difference: in the first category, the data samples used are fixed, whereas in the second category, the data samples used shall be carefully selected in order to achieve a satisfactory accuracy with a high probability. By using the above two-step framework, NTD can be scaled up for large-scale problems, and the error is governed by the quality of the LRA in Step 1, as stated in Proposition \ref{th:ErrB}. 
 
\section{Unique and Sparse NTD}
\label{sec:sparse}
Tucker decompositions are often criticized for suffering from two major disadvantages: the curse of dimensionality and the lack of uniqueness. The former means that the size of the core tensor increases exponentially  with respect to the order $N$, whereas the latter is due to the fact that unconstrained Tucker decompositions essentially only estimate a subspace of each mode. In this section, we discuss how nonnegativity can help  overcome these two limitations of Tucker decompositions, particularly by incorporating sparsity. To our best knowledge, although several NTD algorithms have been developed \cite{NMF-book}, a theoretical analysis of the uniqueness of NTD is still missing.

\subsection{Uniqueness of NTD}
The following notation will be used in the uniqueness analysis:

{\bf Nonnegative Rank: } The nonnegative rank of a nonnegative matrix \mat{Y}, i.e., $\prank{\mat{Y}}$, is equal to the minimal number of $R$ such that $\mat{Y}=\mat{A}\trans{\mat{B}}$, where $\mat{A}\in\Real^{M\times R}_+$ and $\mat{B}\in\Real^{N\times R}_+$. Obviously, $\rank{\mat{Y}}\le\prank{\mat{Y}}$.

{\bf Multilinear rank and nonnegative multilinear rank: } The vector $\mat{r}=(R_1, R_2, \ldots, R_N)$ is called the multilinear rank of \tensor{Y}, where $R_n\defeq\rank{\tenmat{Y}}, \forall n$. The vector $\mats[+]{r}=(R_1, R_2, \ldots, R_N)$ is called the nonnegative multilinear rank of a nonnegative tensor \tensor{Y}, if $R_n\defeq\prank{\tenmat{Y}}$.

{\bf Essential uniqueness: }  We call the NTD $\tensor{Y}=\tensor{G}\times_{n\in\Set{I}_N}\matn{A}$ essentially unique, if $\matn{A}=\mat{\hat{A}}\matn{P}\matn{D}$, $\forall n$, holds for any other NTD $\tensor{Y}=\tensor{\hat{G}}\times_{n\in\Set{I}_N}\matn{\hat{A}}$, where \matn{P} is a permutation matrix, and \matn{D} is a nonnegative diagonal matrix. (On the basis of relationship between NTD and NMF described in \eqref{eq:NMF}--\eqref{eq:TuckerNMF}, the definition of the essential uniqueness of NMF can be obtained.)

Below, we suppose that $\tensor{Y}=\tensor{G}\times_{n\in\Set{I}_N}\matn{A}$ is the NTD of \tensor{Y} with the nonnegative multilinear rank $\mat{r}_+=(R_1, R_2,\ldots, R_N)$, i.e., $\matn{A}\in\Real_+^{I_n\times R_n}$, $R_n=\prank{\tenmat{Y}}$. First, we have the following:

\begin{theorem}
\label{th:prank_ineq}
For any $n\in\Set{I}_N$, $R_n\le R_{\breve{n}}=\prod_{p\neq n}R_p$ holds.
\end{theorem}
\begin{IEEEproof}
Note that $\tenmat{Y}=\matn{A}\tenmat{G}\trans{\matn{R}{}}$, where $\tenmat{G}\in\Real_+^{R_n\times R_{\breve{n}}}$. If there exists $n$ such that $R_n>R_{\breve{n}}$, we simply let $\matn{\tilde{A}}=\matn{A}\tenmat{G}$, $\tenmat{\tilde{G}}=\matI$, and \matn[p]{\tilde{A}}=\matn[p]{A} for all $p\neq n$. Then, $\tensor{Y}=\tensor{\tilde{G}}\times_{n\in\Set{I}_N}\matn{\tilde{A}}$ forms another NTD of \tensor{Y} with $\prank{\tenmat{{Y}}}\le R_{\breve{n}}<R_n$, which contradicts the assumption of $\prank{\tenmat{Y}}=R_n$.  
\end{IEEEproof}

\begin{corollary}
\label{corr:UNTD_Rn}
Let $R_n=\max (R_1,R_2,\ldots,R_N)$, and the NTD $\tensor{Y}=\tensor{G}\times_{n\in\Set{I}_N}\matn{A}$ is essentially unique. If $s(\tensor{G})<1-{1}/{R_n}$, then $R_n < R_{\breve{n}}=\prod_{p\neq n}R_p$.
\end{corollary}
In Corollary \ref{corr:UNTD_Rn}, the condition $s(\tensor{G})<1-{1}/{R_n}$ means that \tenmat{G} is not a trivial matrix that is a product of a permutation matrix and a nonnegative scaling matrix. Following the proof of Proposition \ref{th:prank_ineq}, the proof of Corollary \ref{corr:UNTD_Rn} is obvious.

\begin{theorem}
\label{eq:UNTDnece}
If the NTD $\tensor{Y}=\tensor{G}\times_{n\in\Set{I}_N}\matn{A}$ is essentially unique, then $\tenmat{\tilde{Y}}=\matn{A}\matn{G}$ is the unique NMF of matrix \tenmat{\tilde{Y}}  for all $n\in\Set{I}_N$.
\end{theorem}
\begin{IEEEproof}
Suppose that there exists  $p\in\Set{I}_N$ and a non-trivial matrix \mat{Q} such that $\tenmat[p]{\tilde{Y}}=(\matn[p]{A}\mat{Q})(\mat{Q}{}^{-1}\tenmat[p]{G})$ is another NMF of \tenmat[p]{\tilde{Y}}; then, $\tensor{Y}=\tensor{\tilde{G}}\times_{n\in\Set{I}_N}\matn{\tilde{A}}$ and $\tensor{Y}=\tensor{G}\times_{n\in\Set{I}_N}\matn{A}$ are two different NTDs of \tensor{Y}, with $\matn[p]{\tilde{A}}= \matn[p]{A}\mat{Q}$, \matn{\tilde{A}}=\matn{A} for $n\neq p$, and $\tenmat{\tilde{G}}=\mat{Q}^{-1}\tenmat{G}$. This contradicts the assumption that the NTD of \tensor{Y} is essentially unique.
\end{IEEEproof}

\begin{figure*}[!t]
\centering
\includegraphics[width=.9\linewidth,height=0.25\linewidth]{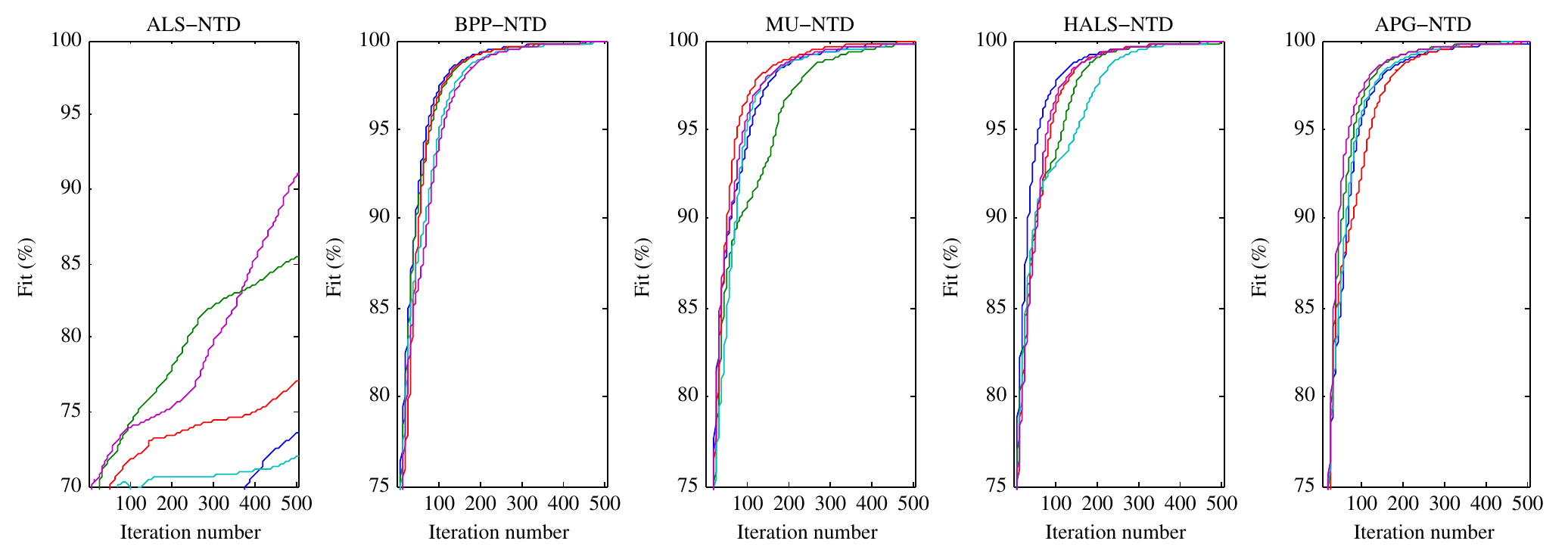}
\caption{The evolution of the Fit values versus the iteration number of each algorithm in five runs. In each run, all of the algorithms started from the same initial settings. The data tensor was generated by using $\tensor{Y}=\tensor{G}\times_{n\in\Set{I}_4}\matn{A}+\tensor{N}$, where the entries of \tensor{G} and \matn{A} were drawn from i.i.d. exponential distributions with $s(\matn{A})=s(\tensor{G})=0.6$. The entries of the noise \tensor{N} were drawn from a standard Gaussian distribution with an SNR of 0 dB.}
\label{fig:convg}
\end{figure*}

\begin{theorem}
\label{th:UniPNTD}
For the population NTD $\tensor{Y}=\tensor{G}\times_{n\neq N}\matn{A}$, if \tenmat[N]{Y} has an essentially unique NMF with the positive rank equal to $\prod_{p\neq N}R_p$, then the population NTD of \tensor{Y} is essentially unique.
\end{theorem}
\begin{IEEEproof}
As \tenmat[N]{Y} has an essentially unique NMF and $\tenmat[N]{Y}=\tenmat[N]{G}\trans{\matn[N]{R}{}}$, where $\matn[N]{R}=\bigkkp_{n\neq N}\matn{A}$, both \tenmat[N]{G} and \matn[N]{R} can actually be essentially uniquely estimated. Without loss of generality, we suppose that 
\begin{equation}
\label{eq:uniRpAp} \notag
\matn[N]{\hat{R}}=\matn[N]{R}\mat{PD}=\left(\bigkkp\nolimits_{n\neq N}\matn{A}\right)\mat{{P}}\mat{{D}},
\end{equation}
where \matn[N]{\hat{R}} is an estimate of \matn[N]{R},  \mat{{P}} is a permutation matrix, and \mat{{D}} is a nonnegative diagonal matrix. In other words,
\begin{equation}
\label{eq:Rn2An}
\matn[N]{\hat{R}}=\bigkkp\nolimits_{n\neq N}\matn{A}\matn{P}\matn{D}\defeq\bigkkp\nolimits_{n\neq N}\matn{\hat{A}},
\end{equation}
where  $\matn{\hat{A}}\defeq\matn{A}\matn{P}\matn{D}$, \matn{P} and \matn{D} are permutation matrices and diagonal matrices such that $\mat{{P}}=\bigkkp_{n\neq N}\matn{P}$ and $\mat{{D}}=\bigkkp_{n\neq N}\matn[n]{D}$. Below, we only need to show that $\matn{\hat{A}}$ can be essentially uniquely estimated from \matn[N]{\hat{R}}, which in turn results in the essential uniqueness of \matn{A}.

Motivated by the method proposed in \cite{kkpapprox}, we appropriately arrange the elements of \matn[N]{\hat{R}} and reshape it to form a tensor $\tensor{\hat{R}}$ such that
\begin{equation}
\begin{split}
\tensor{\hat{R}}=  vec(\matn[1]{\hat{A}}) & \outerp \cdots \outerp vec(\matn[p-1]{\hat{A}}) \\
& \outerp vec(\matn[p+1]{\hat{A}}) \cdots \outerp vec(\matn[N]{\hat{A}}),
\end{split}
\end{equation}
which means that $\tensor{\hat{R}}$ is a rank-one tensor and \matn{\hat{A}} can be uniquely estimated from $\tensor{\hat{R}}$. This ends the proof.
\end{IEEEproof}

\begin{corollary}
\label{corr:UNTD}
Let $\tensor{Y}=\tensor{G}\times_{n\in\Set{I}_N}\matn{A}$ be the NTD of \tensor{Y}. If \tenmat{Y} has an essentially unique NMF with the positive rank of $R_n$ for all $n\in\Set{I}_N$, then the NTD of \tensor{Y} is unique.
\end{corollary}

Corollary \ref{corr:UNTD} describes a special case in which the NTD of a high-order tensor can be achieved by solving $N$ independent NMF subproblems, which avoids the nonnegative alternating least squares with respect to $N+1$ factors and has been realized in \cite{TSP-lraNMF,SPM_NMFNTD}. Furthermore, from the proof of Proposition \ref{th:UniPNTD}, we know that the factors can be essentially uniquely recovered from their Kronecker products. This motivates us to extend the idea of mode reduction (reshaping) proposed in \cite{MRCPD} to NTD; that is, the NTD of an $N$th-order tensor can be implemented by performing NTD on a 3rd-order tensor that is obtained by reshaping the original $N$th-order tensor,  followed by a Kronecker product approximation procedure. Once the 3rd-order tensor has an essentially unique NTD (e.g., all of its three unfolding matrices have a unique NMF), the original $N$th-order tensor also does.  

From the above analysis, the uniqueness of NTD has a very close relation with the uniqueness of NMF. So far, there have been many results on the uniqueness of NMF (see \cite{GillisSUNMF,Laurberg2007a,HuangNMF2013} for a comprehensive review), most of which are based on the sparsity of factor matrices. Among them, the pure-source-dominant condition \cite{nLCAIVM2010, TNN-MVCNMF}, which means for each signal there exists at least one instant at which only this signal is active or strongly dominant, is one of the most popular uniqueness conditions for NMF  \cite{nLCAIVM2010, TNN-MVCNMF, SepNMF_LP,FastSepNMF}:
\begin{theorem}
The NMF of \mat{Y}=\mat{A}\trans{\mat{B}} is essentially unique if \mat{B} satisfies the pure-source-dominant condition, i.e., $\mat{B}=\mat{\Pi}\trans{\begin{bmatrix}
\matI & \trans{\mats[0]{B}}
\end{bmatrix}}\mat{P}\mat{D}$, where \mat{D} is a diagonal scaling matrix, \mat{P} and \mat{\Pi} are permutation matrices, and \matI{} is the identity matrix. 
\end{theorem}

The pure-source-dominant condition is also studied in terms of separable NMF, which gained popularity very recently, as it proved to be highly scalable and its representative applications include  topic discovery and  the clustering analysis  of large-scale datasets \cite{SepNMF_LP,FastSepNMF}. If we replace the unique NMF in Proposition \ref{th:UniPNTD} and Corollary \ref{corr:UNTD} with the above pure-source-dominant condition, we obtain the corresponding uniqueness conditions for NTD.

Note that the pure-source-dominant condition essentially requires that at least one factor matrix of NMF should be sufficiently sparse. In NTD, it requires that the core tensor or all component matrices are sufficiently sparse. In fact, sparsity is not only a key factor of the uniqueness of NTD, it also reflects the learning-parts ability of NTD, as many zeros often exist in the factors. Below we focus on sparse NTD. 

\subsection{NTD with Sparse Core Tensors}
 A very sparse core tensor is of particular importance for NTD:  not only does it partially break the curse of dimensionality, as it only keeps the most significant connections between the components in different modes, it also improves the uniqueness feature of the results. In fact, in the ideal case where  \tensor{G} is very sparse such that \tensor{G} is all-zero except $g_{i,i,\ldots,i}>0$ for $i=1,2,\ldots,\min(R_1,R_2,\ldots, R_N)$,  NTD is essentially reduced to  nonnegative polyadic decomposition (NPD) \cite{NMF-book,SPM_NMFNTD}, which is essentially unique under very mild conditions (even if \matn{A} is not sparse)\cite{Sidiropoulos2000}.  All of these facts suggest that a very sparse core tensor is quite useful in practice. Below, we focus on how to improve the sparsity of the core tensor by imposing suitable constraints, which can also be applied to improve the sparsity of component matrices similarly.

One popular approach is to use the $l_1$ penalty to improve the sparsity of \tensor{G}, leading to $D_{\text{SNTD}}= D_{\text{NTD}}+\lambda\frob[1]{\tensor{G}}$. As \tensor{G} is nonnegative, we have
\begin{equation}
\label{eq:SNTD_G_grad}
\fpartial{D_\text{SNTD}}{\tensor{G}}=\fpartial{D_\text{NTD}}{\tensor{G}}+\lambda,
\end{equation}
where $\fpartial{D_\text{NTD}}{\tensor{G}}$ is given in \eqref{eq:LRAgrad}. Hence, the aforementioned algorithms can be applied directly.

Another approach is to add the Frobenius norm penalty on \matn{A} such that $D_{\text{SNTD}}= D_{\text{NTD}}+\sum_n\frac{\lambda_n}{2}\frob[F]{\matn{A}}^2$, which generally leads to denser factor matrices \matn{A} but a more sparse core tensor \tensor{G}. In such a case, each subproblem with respect to \matn{A} is strictly convex and equivalent to applying  Tikhonov regularization.

\subsection{NTD with Sparse Mode Matrices}
We consider the partial NTD model in \eqref{eq:PNTD}--\eqref{eq:PNTDMat}. Notice that the basis matrix $\bigkkp_{{n\neq N}}\matn{A}$ has a special Kronecker product structure that is not possessed by NMF. Below, we show that such a Kronecker product structure will substantially improve the sparsity of the basis matrix.

{\bf Lemma 1 \cite{SPM_NMFNTD}:}
Let $\mat{a}\in\Real^{M\times 1}$ and $\mat{b}\in\Real^{N\times 1}$. Then, $z_{\mat{a}\krp\mat{b}}=Nz_{\mat{a}}+Mz_{\mat{b}}-z_{\mat{a}}z_{\mat{b}}$ and $s_{\mat{a}\krp\mat{b}}=s_{\mat{a}}+s_{\mat{b}}-s_{\mat{a}}s_{\mat{b}}$, which means  $s_{\mat{a}\krp\mat{b}}\ge \max(s_{\mat{a}},s_{\mat{b}})$. 
\begin{theorem}
\label{th:kkpsp}
Let $\matn{A}\in\Real^{I_n\times R_n}$, $n=1,2$. Then,  $s_{{\matn[1]{A}}\kkp\matn[2]{A}}=s_{\matn[1]{A}}+s_{\matn[2]{A}}-s_{\matn[1]{A}}s_{\matn[2]{A}}\ge\max(s_{\matn[1]{A}},s_{\matn[2]{A}})$.
\end{theorem}
\begin{IEEEproof}
Let $\mat{K}=\matn[1]{A}\kkp\matn[2]{A}$. There exists a rearrangement of \mat{K}, denoted as $\mats[\text{R}]{K}$, satisfying $\mats[\text{R}]{K}=vec(\matn[1]{A})\trans{vec(\matn[2]{A})}$ (see \cite{kkpapprox}), or equivalently, 
\begin{equation}
\label{eq:kkpvec}
vec(\mats[\text{R}]{K})=\left[vec(\matn[2]{A})\right]\krp \left[vec(\matn[1]{A})\right].
\end{equation} 
As the arrangement and vectorization operators do not change the values of the entries, from \eqref{eq:kkpvec} and Lemma 1, we have
\begin{equation}
\label{eq:kkpz}
z_{\matn[1]{A}\kkp\matn[2]{A}}=I_2R_2z_{\matn[1]{A}}+I_1R_2z_{\matn[2]{A}}-z_{\matn[1]{A}}z_{\matn[2]{A}},
\end{equation}
and the rest of the proof is obvious.
\end{IEEEproof}
From Proposition \ref{th:kkpsp}, NTD generally is able to provide more sparse basis matrices than NMF. This sparsity stems from the sparsity of each factor matrix and is further enhanced by the Kronecker product operators.

 \begin{figure}[!t]
\centerline{
 \subfloat[Fits vs. noise levels]{
    \includegraphics[width=.9\linewidth]{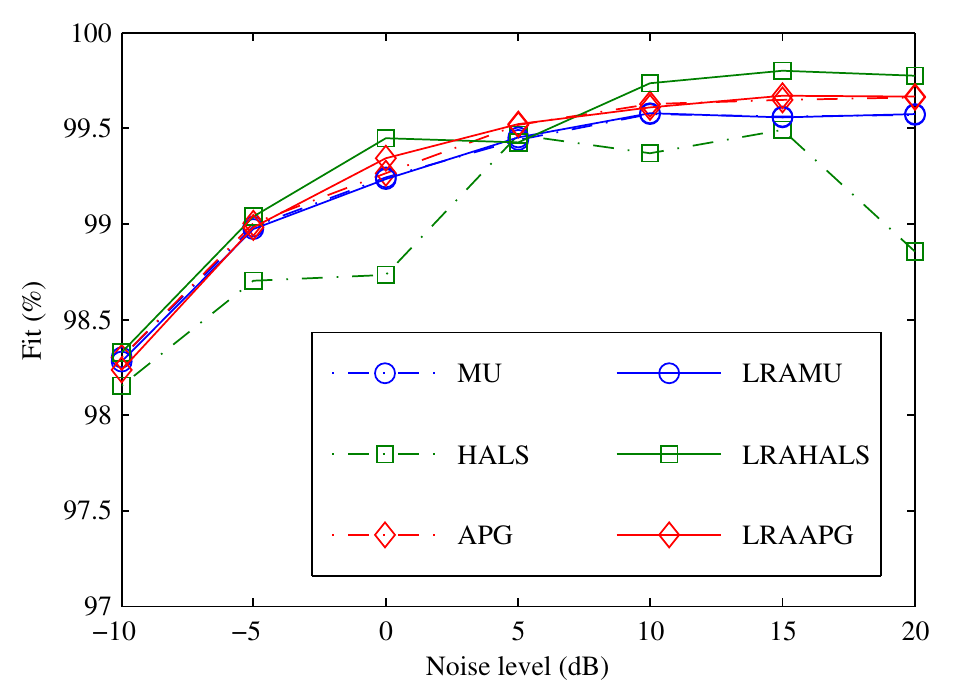}} 
}
\centerline{
 \subfloat[Elapsed times for each algorithm]{
    \includegraphics[width=.9\linewidth]{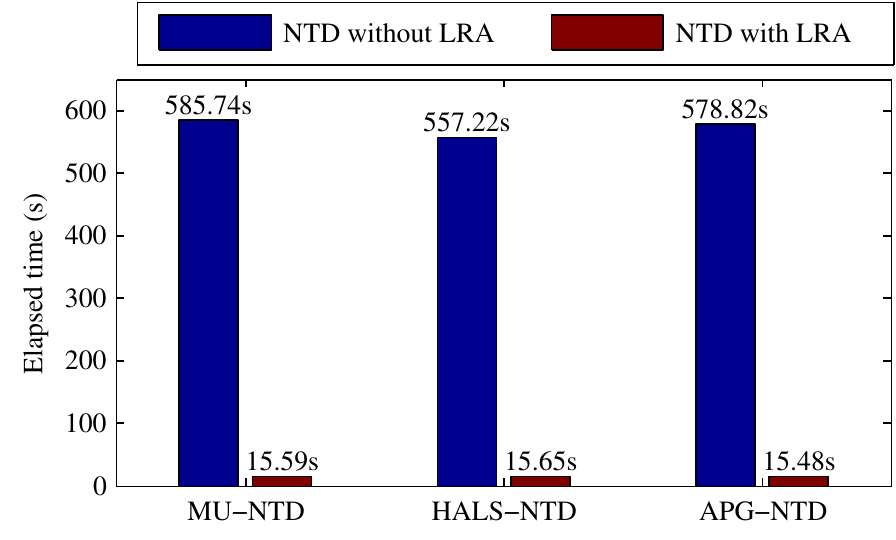}} 
}
\caption{Comparison between the NTD algorithms with or without LRA under different levels of noise. The NTD algorithms based on LRA are more robust to noise and are significantly faster than those without LRA.}
\label{fig:noiLevels}
\end{figure}

\begin{figure}[!t]
\centering
\includegraphics[width=.9\linewidth]{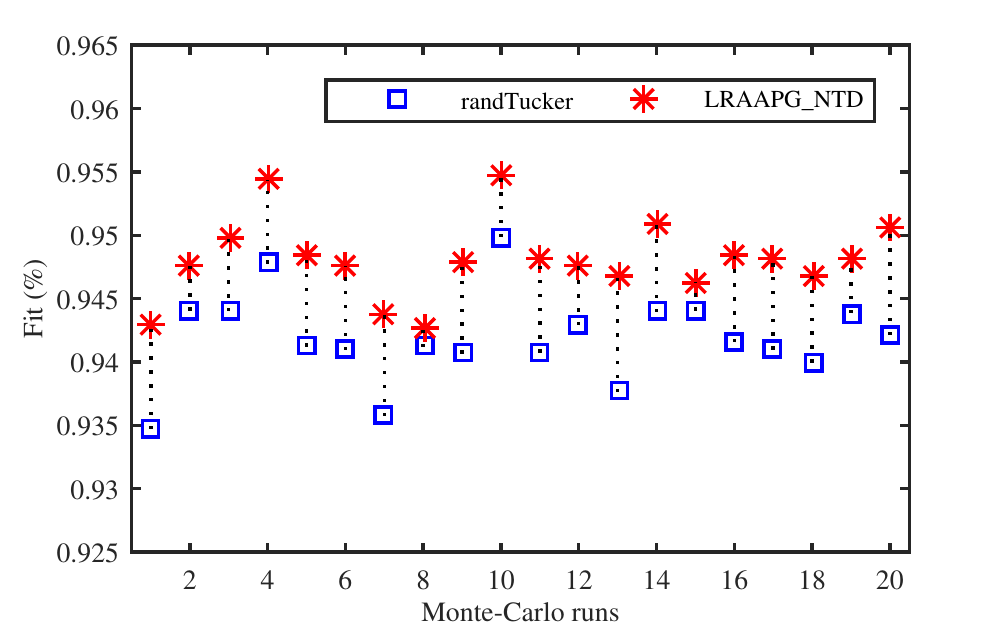}
\caption{Illustration of how the performance of the LRA affected the final results over 20 Monte Carlo runs , where randTucker  \cite{bigTensor2014} was used to compress the noisy data tensor. Roughly speaking, a more accurate LRA leads to better  results.}
\label{fig:LRAeffect}
\end{figure}

\begin{figure}[!t]
\centerline{
 \subfloat[Fits vs. sparsity levels]{
    \includegraphics[width=.9\linewidth]{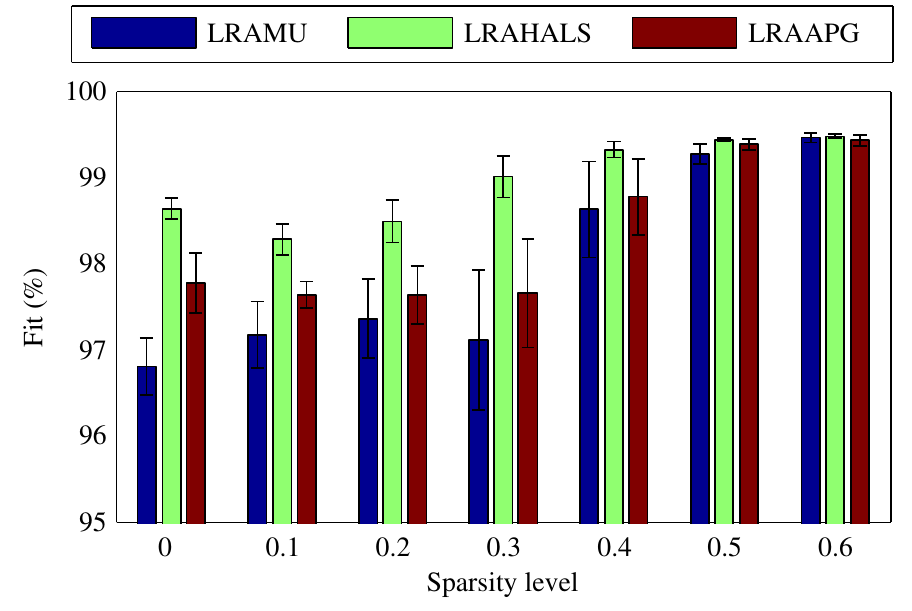}} 
}
\centerline{
 \subfloat[mSIRs vs. sparsity levels]{
    \includegraphics[width=.9\linewidth]{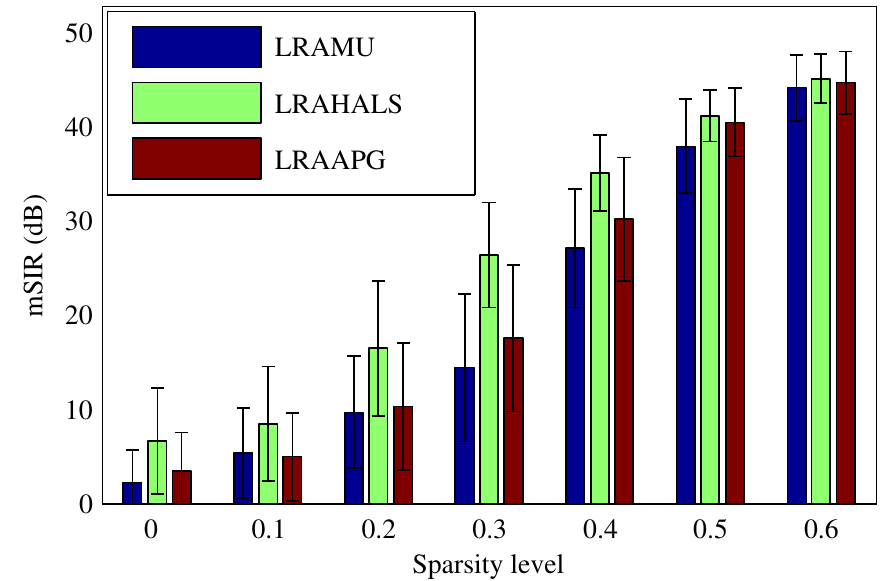}} 
}
\caption{The Fit and mSIR values versus the sparsity level of factors averaged over 10 Monte Carlo runs, where $s(\matn{A})=s(\tensor{G})=p$,  $p\in\set{0,0.1,\ldots,0.6}$, and the additive Gaussian noise was 0 dB. It can be seen that if the factors were very sparse, all the LRA-NTD algorithms were able to recover the true components in a very high probability. }
\label{fig:sparsity}
\end{figure}

\section{Simulation Results and Applications}
\label{sec:simu}
In this section, the performance of the proposed algorithms is demonstrated by using both synthetic and real-world data. All  simulations were performed on a computer with an i7CPU at 3.33GHz and 24GB memory, running Windows 7. The MATLAB codes of the proposed algorithms are available at  \url{http://bsp.brain.riken.jp/~zhougx}.
\subsection{Simulations Using Synthetic Data}
{\emph{Basic Settings: }} The data tensor \tensor{Y} was generated by using $\tensor{Y}=\tensor{G}\times_{n\in\Set{I}_4}\matn{A}+\tensor{N}$, where the elements of the component matrices and the core tensor were drawn from independent exponential distributions with the mean parameter 10. The entries of the additive noise term \tensor{N} were drawn from independent Gaussian distributions. In all simulations  using synthetic data, we set $I_n=100$, $n=1,2,3,4$, and the nonnegative multilinear rank $\mat{r}_+=(5, 6, 7, 8)$. To generate the sparse core tensor and component matrices, some entries that were uniformly sampled  from each matrix/tensor were set to  be zero to meet the specified sparsity. We used two performance indices to measure the approximation accuracy. The first one is the Fit index, which measures the fitting error between the data tensor\footnote{For the simulations using synthetic data, the tensor \tensor{Y} in Fit(\tensor{Y},\tensor{\hat{Y}}) is the latent noise-free tensor; otherwise, for real-world data, \tensor{Y} is the data tensor.} \tensor{Y} and its estimate \tensor{\hat{Y}}:
\begin{equation}
\label{eq:fit} \notag
\text{Fit}(\tensor{Y},\tensor{\hat{Y}})=\left(1-{\frob{\tensor{Y}-\tensor{\hat{Y}}}}/{\frob{\tensor{Y}}}\right)\times 100\%.
\end{equation}
Another performance index, the mean signal-to-interference ratio (mSIR, dB), measures how well the recovered components match the true components:
\begin{equation}
\label{eq:mSIR} 
\text{mSIR}=\frac{1}{\sum_n R_n}\sum_{n=1}^{N}\sum_{r_n=1}^{R_n}20\log_{10}{\frac{\frob[2]{\matn{a}_{r_n}}}{\frob[2]{\matn{a}_{r_n}-\matn{\hat{a}}_{r_n}}}},
\end{equation}
where $\matn{\hat{a}}_{r_n}$ is an estimate of $\matn{a}_{r_n}$, both of which are normalized to have zero-mean and unit variance. 

On the basis of the NLS solvers we introduced in Section \ref{sec:LRANTD}, we implemented five NTD algorithms: MU-NTD, HALS-NTD, BPP-NTD, APG-NTD, and ALS-NTD, each of which has  versions  with or without LRA. In our implementation of BPP-NTD, we have borrowed code from  BPP-NMF \cite{NMF_BPP} to solve each NLS subproblem.

1. Convergence speed of different update rules. The maximum iteration number for each NLS subproblem was 20, and the total number of iterations was 500 for all algorithms. In this comparison, we directly ran each algorithm on the observation data  without the LRA procedure to compare the performance between the suggested update rules, and the evolution of the Fit values versus the iteration number is shown in \figurename \ref{fig:convg} by using five different initial values. It can be seen that the ALS-NTD algorithm was quite sensitive to the initial values, mainly because it involves the computation of the inverse of probably ill-conditioned matrices during the iterations. (This issue was also observed in the BPP-NTD algorithm \cite{Guan2012}, although it seems to be not as serious as in  ALS-NTD. It seems that  APG-NTD is less sensitive to the initial values, whereas  HALS-NTD often provides a higher accuracy and faster convergence speed. Hence, in the comparisons below, we focused on the other three more stable algorithms). Except the ALS-NTD algorithm, all  algorithms converged consistently.

2. Comparison between the NTD algorithms with or without LRA for different levels of noise. For each algorithm, the stopping criterion was $\frob{\matn{A}_{iter+1}-\matn{A}_{iter}}^2<10^{-6}$.  All of the LRA-based NTD algorithms used  HOSVD \cite{HOSVD2000} to obtain the LRA of the noisy observation data. Their performance averaged over 10 Monte Carlo runs is shown in \figurename \ref{fig:noiLevels}. From  \figurename \ref{fig:noiLevels}(a), we can see that the LRA-based NTD algorithms are often more robust than those without LRA. We guess this is mainly because the NTD algorithms without LRA are more sensitive to the initial values for noisy data due to the nonnegative projection during iterations. In other words, LRA is quite helpful to reduce the noise and consequently improve the robustness of the NTD algorithms. Moreover, just as expected, the LRA-based NTD algorithms are significantly faster than those without LRA, as shown in \figurename \ref{fig:noiLevels}(b).

We also investigated how LRA will affect the final results of NTD. In this experiment, the randTucker algorithm proposed in \cite{bigTensor2014} was used to compress the data tensor (contaminated by Gaussian noise with SNR = 20 dB). Then, LRAAPG was applied to perform NTD by using exactly the same initial settings.  \figurename \ref{fig:LRAeffect} shows the results of 20 Monte Carlo runs, which demonstrated that a more accurate LRA roughly led to better performance for NTD. Note also  that the final Fit values were often higher than those of LRA (the Fit was evaluated using the noise-free tensor instead of the observation tensor), suggesting the nonnegativity constraints may help to remove noise and improve the estimation accuracy.

3. Investigation of how sparsity affects the essential uniqueness of NTD by applying the developed algorithms to the data whose factors were of different levels of sparsity. In this simulation,  we set $s(\matn{A})=s(\tensor{G})=p$, $p\in\set{0,0.1,\ldots,0.6}$ each time. In each run, 0 dB Gaussian noise was added to the data tensor data, and the LRA procedure of all the algorithms was again performed by using  HOSVD \cite{HOSVD2000}. The average performance over 10 Monte Carlo runs is shown in \figurename \ref{fig:sparsity}. From the figure, when the sparsity of the factors (including the core tensor) were higher than 0.3, the mSIR values were generally higher than 20 dB, which means that the corresponding NTDs in such cases were essentially unique, and the existing NTD algorithms were able to recover the true components with a very high probability. However, if the factors were of low sparsity level, all algorithms not only failed to recover all true components but also achieved lower Fit values. We guess that this was mainly caused by the local convergence of the NTD algorithms. From the simulation results, the sparsity of factors is one key factor in NTD, as it substantially improves the essential uniqueness of NTD, which in turn leads to a better fit to data, as sticking in local minima of the NTD algorithms can be largely avoided.

\begin{figure*}[!th]
\centerline{
    \includegraphics[width=.75\linewidth,height=0.3\linewidth]{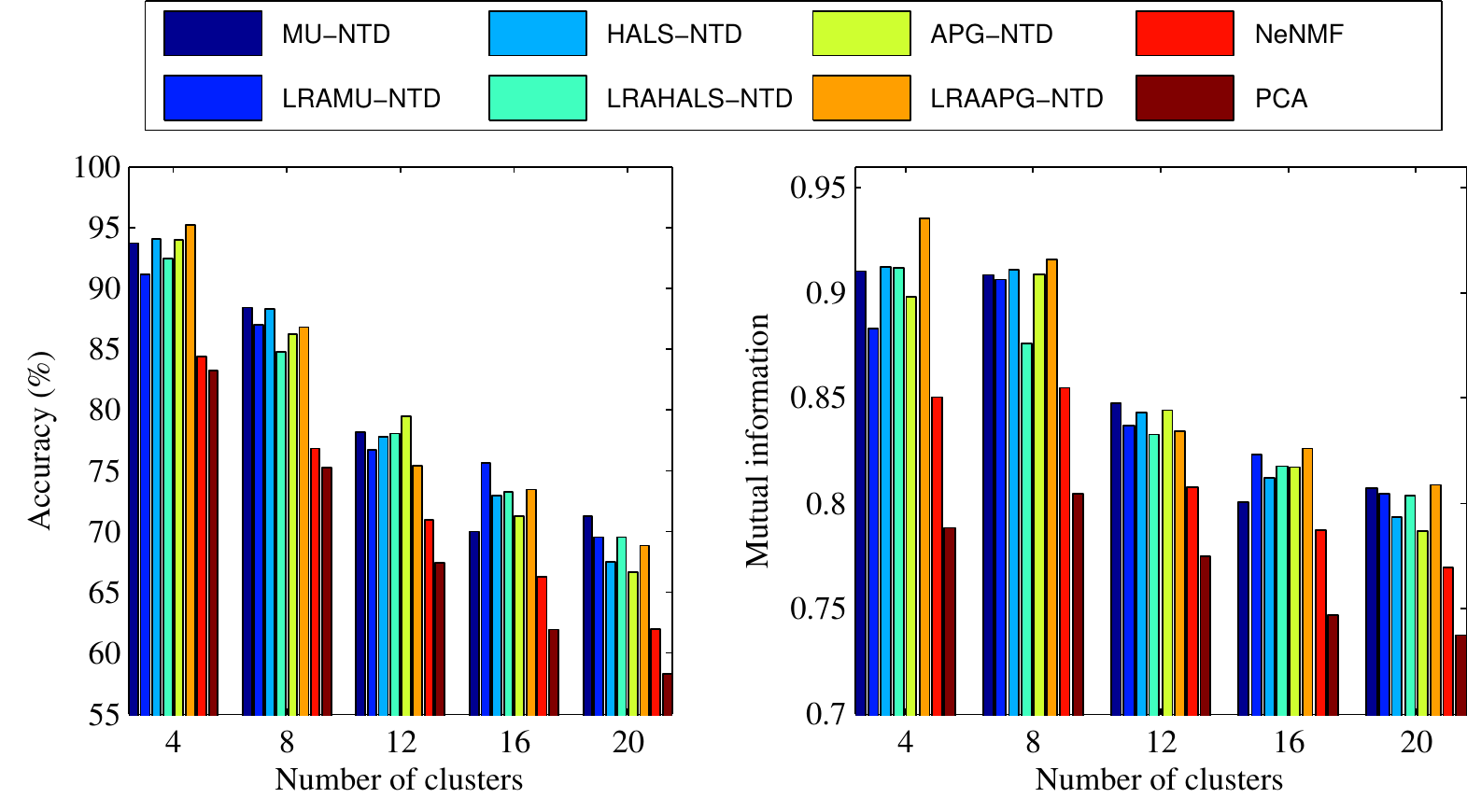}
}
\caption{Performance comparison between the proposed algorithms and the standard NMF/PCA algorithms when they were applied to perform the clustering analysis of the COIL100 objects. }
\label{fig:COIL100acci}
\end{figure*}

\subsection{Experiments Using Real-world Data}
\begin{figure}[!th]
\centerline{
 \subfloat[Basis images learned by NMF]{
    \includegraphics[width=.47\linewidth,height=.6\linewidth]{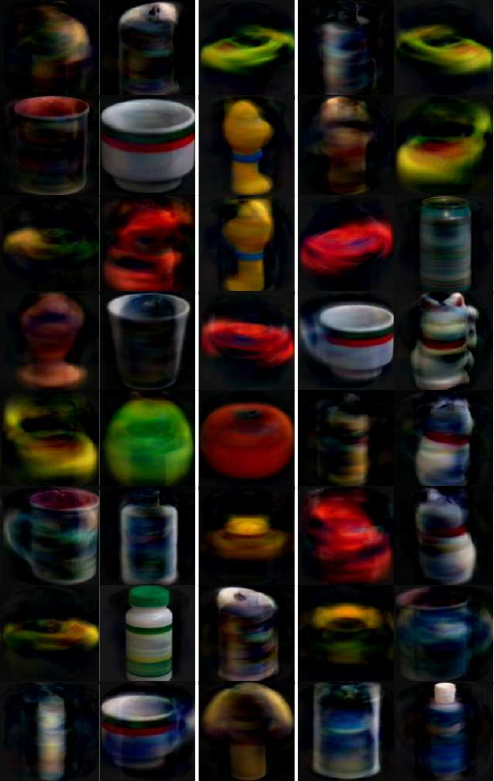}} 
 \subfloat[Bass images learned by NTD]{
    \includegraphics[width=.475\linewidth,height=.6\linewidth]{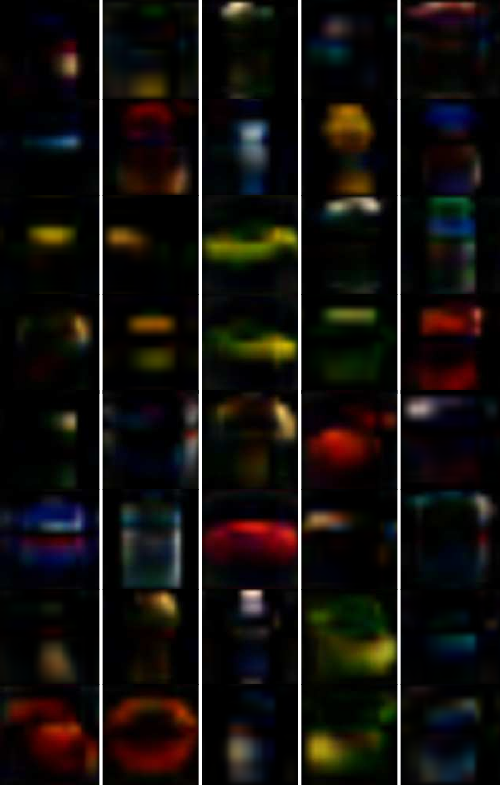}} 
}
\caption{Basis images learned by using  NeNMF and  LRAHALS-NTD  from  20 categories randomly selected from the COIL-100 database. The bases learned by NTD ($\tensor{G}\times_{n\in\Set{I}_3}\matn{A}$) are more sparse than thoese obtained by  NeNMF.}
\label{fig:COIL100_basis}
\end{figure}

{\bf Object clustering.} In this experiment, we applied the proposed NTD algorithms to the clustering analysis of  objects selected from the Columbia Object Image Library (COIL-100).  The COIL-100 database consists of 7,200 images of 100 objects, each of which has 72 images taken from different poses. For simplicity, we only considered the first 20 categories, and  we randomly selected $k$ categories each time to form a data tensor \tensor{Y} of $128\times128\times3\times 72k$. Then, the tensor \tensor{Y} was decomposed by the proposed NTD algorithms as $\tensor{Y}\approx\tensor{G}\times_{n\in\Set{I}_4}\matn{A}$ by empirically setting $R_1=R_2=10, R_3=3$, and $R_4=2k$ denoting the number of features. We used the factor matrix \matn[4]{A} as the features and used the K-means approach to cluster the objects. In each run of the K-means it was repeated 20 times to mitigate the local convergence issue. To show the superiority of the NTD methods in high-dimensional data analysis, we also used the NeNMF method (accelerated by using the LRA technique proposed in \cite{TSP-lraNMF}) and the PCA method to extract the features from the vectorized samples. For each $k=4,8\ldots,20$, we randomly selected 10 different subsets of objects, and the average performance is plotted in \figurename \ref{fig:COIL100acci} and listed in TABLE \ref{tab:COIL20} (for $k=20$), which indicates that  the NTD approaches outperformed  NeNMF and PCA that work on flattened data, and the LRA-based NTD algorithms were significantly faster than the others. From \figurename \ref{fig:COIL100_basis}, we can observe that the NTD approaches extracted a more sparse and local parts-based basis. Moreover, the core tensors obtained by NTDs were generally very sparse, even without imposing additional sparsity constraints. In this example, 25\% of the entries captured more than 99.5\% of the energy of the entire core tensor, which allows for the adoption of efficient sparse representations of it during  storage and computation. Note also that we did not include the existing NTD algorithms in \cite{HONMF:2008:Morup,HALSNTD:Phan2011,NTD-CVPR2007} for comparison because they are much slower than the proposed algorithms, as analyzed in Section IV. For example, for $k=20$ after 100 iterations, the NTD algorithm proposed in \cite{HONMF:2008:Morup} had consumed approximately 2,360 s while only achieving the Fit of 0.66. These algorithms are inefficient because they update the parameters only once in each iteration and without the LRA acceleration.

\begin{table}[!t]
\caption{Performance Comparison When the Algorithms Were Applied to a Clustering Analysis by Using the First 20 Objects of the COIL-100 Objects, Averaged Over 10 Monte Carlo Runs. The NTD Algorithms Achieved a Higher Clustering Accuracy Than the Nenmf Algorithm.}
\label{tab:COIL20}
\centerline{
\begin{tabular}{ r  c  c  c  c }
\hline \hline
 Algorithm & Accuracy & MI & Fit (\%) & Time (s) \\
 \hline
MU-NTD               & $69.7\pm{3.6}$   & 0.80   & 0.75 &  301 \\
LRAMU-NTD        & $69.4\pm{3.8}$   & 0.81   & 0.74 &  22 \\
HALS-NTD           & $68.0\pm{4.3}$    & 0.79   & 0.75 & 304 \\
LRAHALS-NTD    & $70.0\pm{3.6}$   & 0.81   & 0.74 &  33 \\
APG-NTD             & $68.8\pm{4.9}$   & 0.80  & 0.75 &  340 \\
LRAAPG-NTD      & $67.3\pm{3.9}$   & 0.80   & 0.74  & 23 \\
NeNMF                 & $63.3\pm{4.2}$   & 0.77   & 0.77  & 214 \\
  \hline \hline
\end{tabular}
}
\end{table}

\begin{table*}
  \caption{Comparison of Face Recognition Accuracy Achieved by Each Algorithm Using the PIE, ORL, and Yale Databases Averaged Over Five Monte Carlo Runs. In Each Run, $p\times100\%$ Samples of Each Person Were Used as the Training Data, Whereas the Others Were Used for Testing.}
  \label{tab:FaceReco}
  \centering
  \begin{tabular}{ r | c c c  ||  c c c || c c c}
  \hline \hline
  \multirow{2}{*}{Algorithm}
  & \multicolumn{3}{c||}{PIE ($R_3=70$)}   & \multicolumn{3}{c}{ORL  ($R_3=40$)} & \multicolumn{3}{c}{Yale ($R_3=40$)}   \\
  \cline{2-10}
  & $p=$30\%     & $p=$40\%    & $p=$50\% & $p=$30\% & $p=$40\% & $p=$50\% & $p=$30\% & $p=$40\% & $p=$50\%  \\ \hline

MU-NTD			& 90.2    & 94.1    & 94.7    & {\bf84.6}    & 85.4    & 89.9  & 64.7 & 67.6 & 66.7   \\
LRAMU-NTD		& 89.3    & 93.5    & 94.4    & 83.2    & {\bf85.9}    & 90.0  & 64.5 & 68.4 & {\bf67.6}    \\
HALS-NTD		& 90.1    & 93.5    & 94.3    & 84.3    & 85.2    & 89.5    &63.4 & 67.2 & 66.4  \\
LRAHALS-NTD		& 89.9    & 93.4    & 94.9    & 83.7    & 84.7    & {\bf90.3}  & {\bf64.8}  & 67.0 & 64.4    \\
APG-NTD			& 90.2    & 93.6    & 94.2    & 83.1    & 85.1    & 90.2    &63.5 & 66.5 & 67.3 \\
LRAAPG-NTD		& {\bf90.6}    & {\bf94.2}    & {\bf95.0}    & 82.7    & 84.9    & 90.1  & 64.3 & {\bf68.2} & 65.3   \\
\hdashline
mahNMF			& 85.2    & 90.0    & 92.6    & 81.8    & 83.8    & 89.1  & 60.0 & 64.2 & 65.6   \\
NeNMF			& 86.3    & 91.6    & 94.1    & 68.8    & 75.7    & 83.6   &50.0 & 56.8 & 57.8  \\
PCA				& 80.9    & 87.6    & 91.6    & 80.2    & 83.2    & 88.1   & 63.5 & 64.0 & 62.4  \\

  \hline \hline
  \end{tabular}
\end{table*}

\begin{figure*}[!th]
\centerline{
    \includegraphics[width=.9\linewidth]{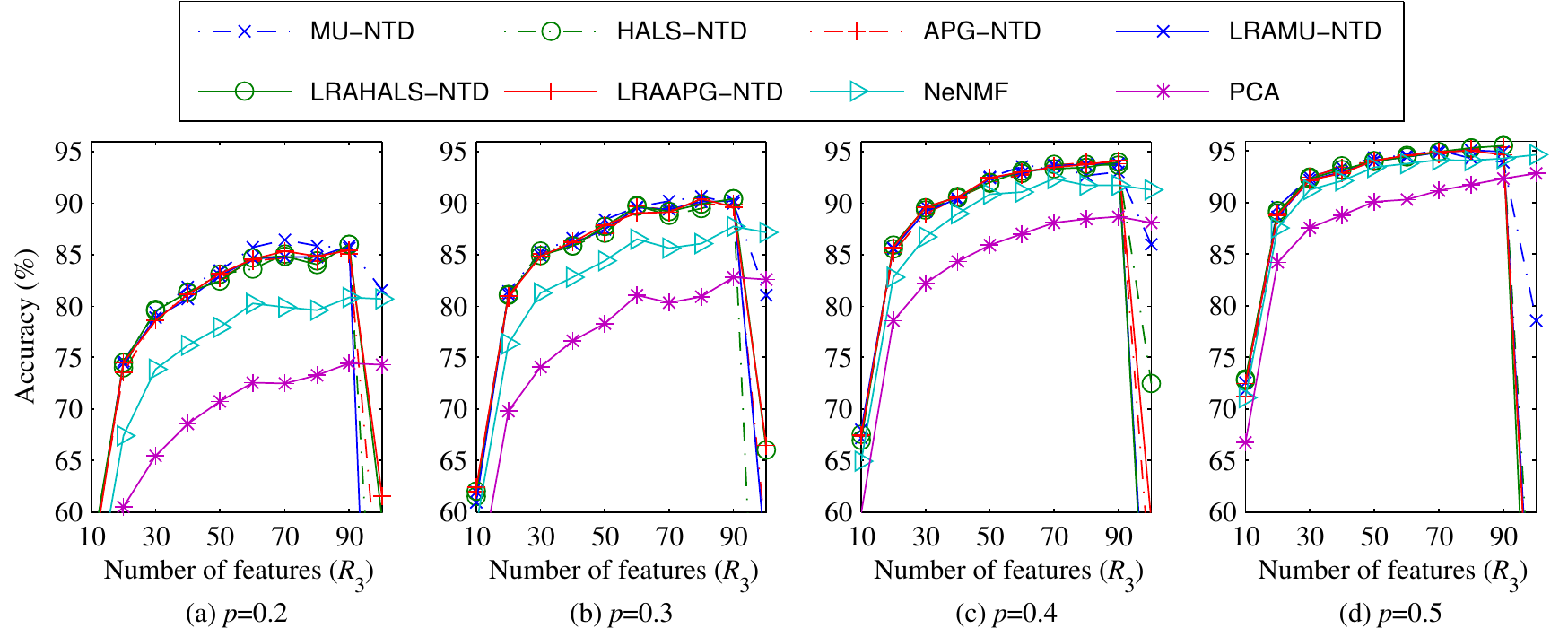}
}
\caption{Face-recognition accuracy  averaged over 5 Monte Carlo runs on the PIE dataset. In each run, $p\times100\%$ of the samples were used for training while the others were used as test samples. The LRAHALS-NTD algorithm was used for feature extraction with varying number of features $R_3$ and fixed $R_1=R_2=10$. }
\label{fig:PIE_R}
\end{figure*}

\begin{figure*}[!th]
\centerline{
 \subfloat[Example face images]{
    \includegraphics[width=.245\linewidth]{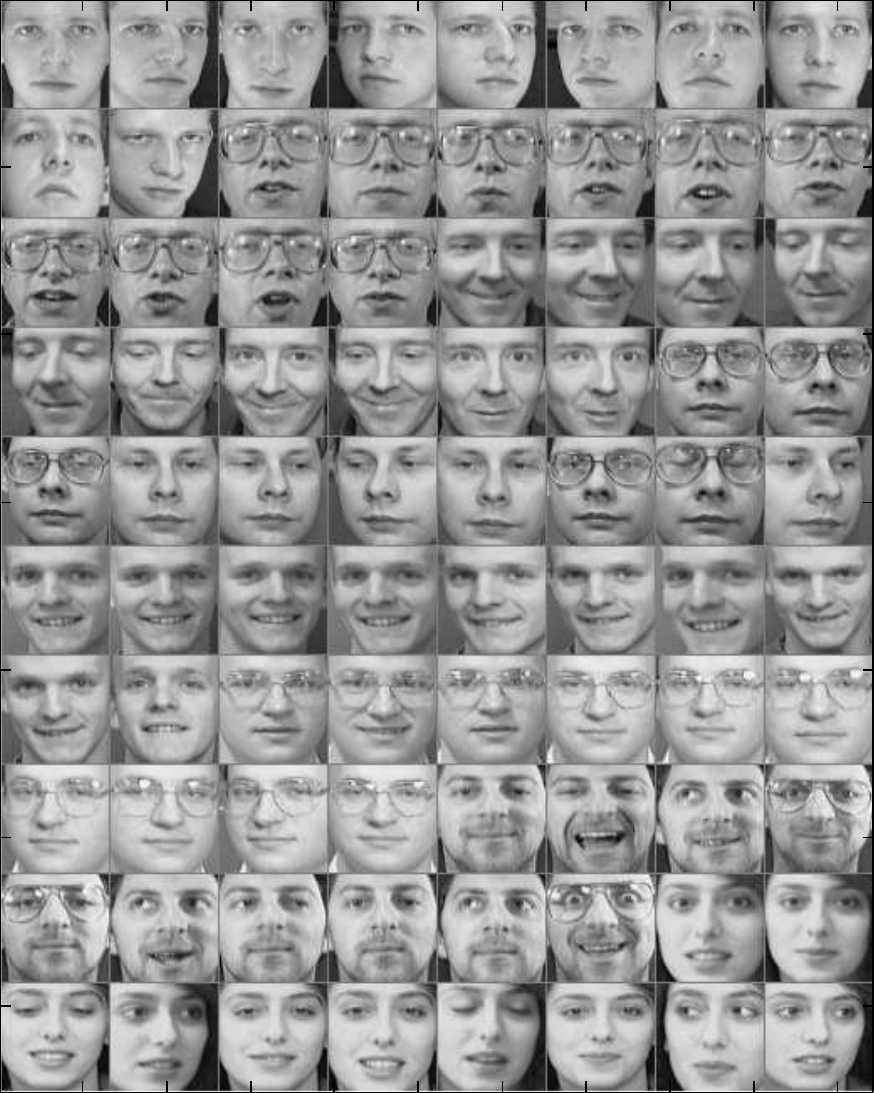}} 
 \subfloat[Basis learned by NeNMF]{
    \includegraphics[width=.245\linewidth]{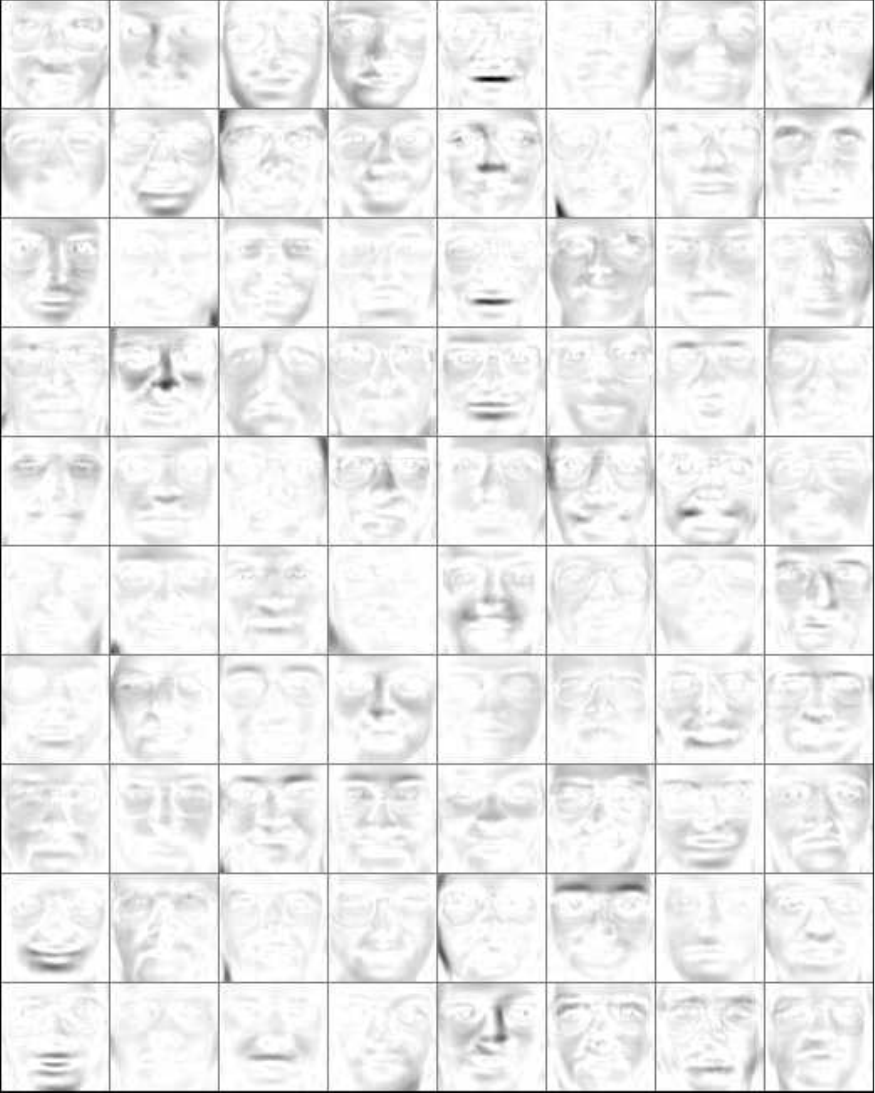}} 
     \subfloat[Basis learned by mahNNF]{
    \includegraphics[width=.245\linewidth]{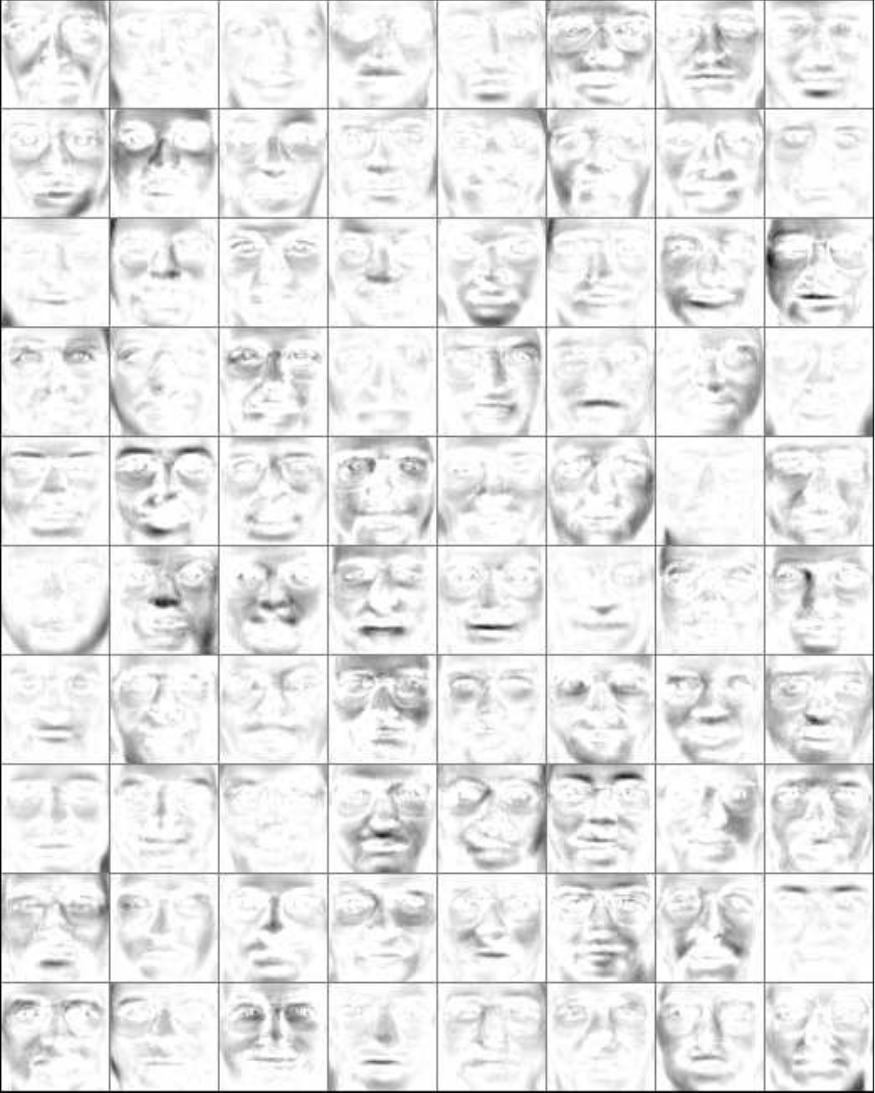}} 
     \subfloat[Basis learned by LRAHALS-NTD]{
    \includegraphics[width=.245\linewidth]{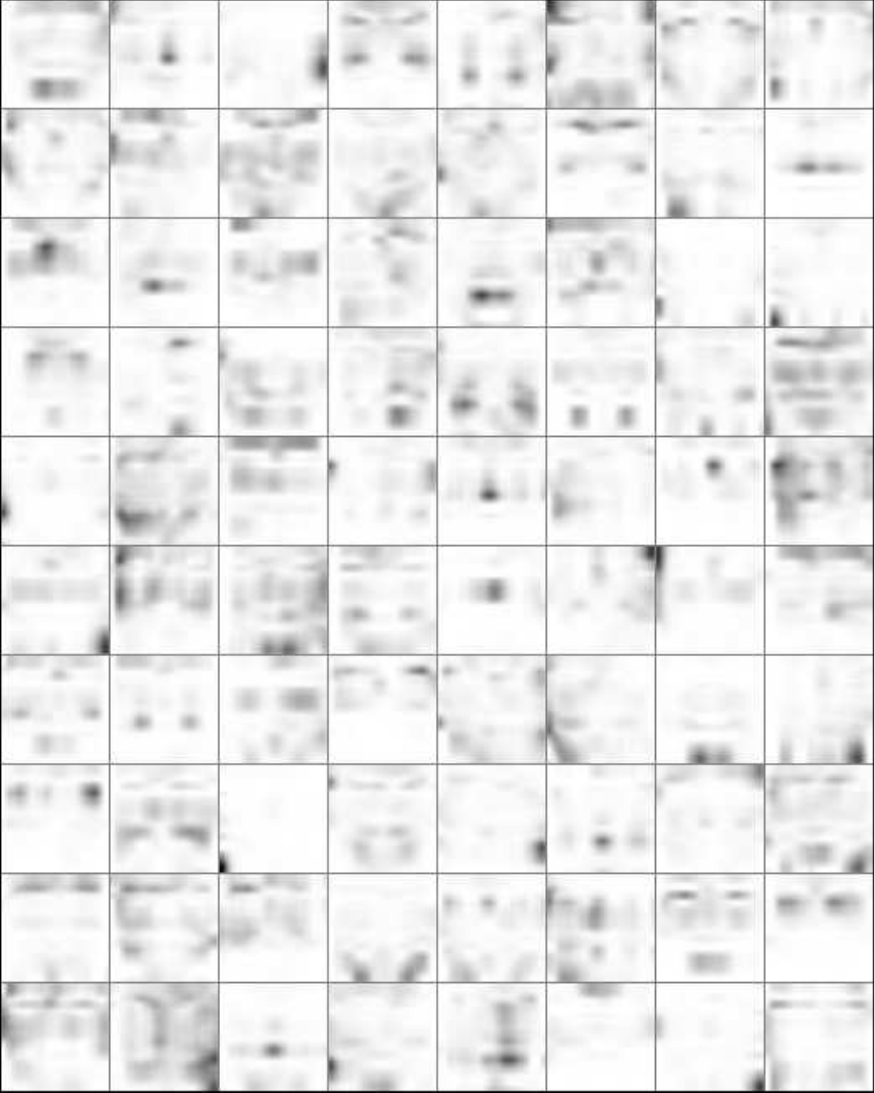}} 
}
\caption{Basis images learned by  NeNMF, mahNMF, and LRAHALS-NTD  from the ORL database. For this data set, whereas the NMF approaches often give global-based representations, LRAHALS-NTD was able to give a localized representation.}
\label{fig:ORLrep}
\end{figure*}

{\bf Human face recognition.} In this experiment, we applied the proposed NTD algorithms to extract features for human face recognition using the PIE (the face images taken at the front pose labeled  c27 were used), ORL, and Yale databases. All of the face images were gray-scaled with a size of $64\times64$.  Each time, we randomly selected $p\times 100\%$ sample images from each person to be the training data, whereas the others were used for testing. In the NTD-based approaches, we decomposed the training data by using the proposed algorithms with empirically setting $R_1=R_2=10$. Then, the unfolding matrix of $\tensor{G}\times_1\matn[1]{A}\times_2\matn[2]{A}$ was used as the basis matrix. For comparison,  mahNMF \cite{mahNMF},  NeNMF, and  PCA were also used to learn the basis matrix from the flattened training data with the same number of features. Once the basis matrix had been learnt from the training data,  the nonnegative projection of a new test sample onto each basis matrix was used as the features for recognition (see \cite{mahNMF}). Finally, the KNN classifier included in MATLAB was used for recognition by using the extracted features, where the distance was measured by their correlations. TABLE \ref{tab:FaceReco} summarizes the recognition accuracy averaged over five Monte Carlo runs. From the table, the NTD algorithms provided a higher accuracy than matrix-factorization-based methods, especially when the amount of training data was relatively small. This phenomenon was also observed in tensor-based discriminate analysis, which shows that the tensor-based methods could considerably alleviate the overfitting problem \cite{STDA_ERP2013}. It can also be seen that the difference between the accuracy obtained by the standard NTD algorithms and that by their LRA-accelerated versions is marginal (in addition, it seemed that the LRA-based versions were more stable). In \figurename \ref{fig:PIE_R}, we illustrate how the number of features, i.e., $R_3$, affected the recognition accuracy by using the PIE database. Basically, a larger value of $R_3$ often led to a higher accuracy, at the cost of a higher computational load. However, once $R_3\ge100$, the performance of the NTD approaches became unstable. We guess this is because NTD is not unique anymore (see Proposition \ref{th:prank_ineq}). NMF was originally developed in order to give a parts-based representation of images and to perform dimensionality reduction in the physical domain. \figurename \ref{fig:ORLrep} shows the basis images learnt by the NeNMF,  mahNMF, and  LRAHALS-NTD algorithms. For the ORL database, it is well known that the NMF algorithms often tend to give global representations rather than parts-based ones, as shown in \figurename \ref{fig:ORLrep}(b) and (c). In contrast, the LRAHALS-NTD algorithm extracted the localized parts of faces without the need to impose any additional sparsity constraints.

\section{Conclusion}
\label{sec:conclusion}
NTD is a powerful tool to analyze multidimensional nonnegative tensor data with the aim of giving a sparse localized parts-based representation of high-dimensional objects. In this paper, we proposed a family of first-order NTD algorithms based on a preceding LRA of the data tensors. The proposed algorithms  use only the first-order information (gradients)  and are free of line search to search update steps (learning rates). The LRA procedure  significantly reduces the computational complexity of the subsequent nonnegative factorization procedure in terms of both time and space and also substantially improves the robustness to noise and the flexibility of the NTD algorithms. Indeed, by incorporating various well-established LRA techniques, the proposed NTD algorithms could be seamlessly implemented to analyze  data contaminated by various types of noise seamlessly. The error bounds on  LRA-based NTD were briefly discussed, and some preliminary results on the essential uniqueness of NTD were provided with a focus on the relationship to the uniqueness of NMF. We discussed how sparsity is able to improve the uniqueness of NTD and to partially alleviate the curse of dimensionality of Tucker decompositions. Simulations justified the efficiency of the proposed LRA-based NTD algorithms and demonstrated their promising applications in clustering analysis.

\bibliographystyle{IEEEtran}
\input{TIP-12015-2014.bbl}


\end{document}